\documentclass[10pt,twocolumn,a4paper]{article}

\usepackage[utf8]{inputenc}
\usepackage{graphicx}
\usepackage{multicol}
\usepackage{amsmath,amssymb}
\usepackage{caption}
\usepackage{subcaption}
\usepackage{booktabs}
\usepackage{hyperref}
\usepackage{geometry}
\usepackage{enumitem}
\usepackage{titlesec}
\usepackage{fancyhdr}
\usepackage{appendix}
\usepackage[numbers]{natbib}
\usepackage{authblk}
\usepackage{multirow}
\usepackage{amssymb}
\usepackage{textcomp}
\usepackage{wrapfig}
\usepackage{tablefootnote}
\usepackage{footnotehyper}
\usepackage[T1]{fontenc}
\usepackage{xcolor}
\usepackage{listings}

\lstdefinelanguage{js}{
  keywords={break, case, catch, continue, debugger, default, delete, do, else, finally, for, function, if, in, instanceof, new, return, switch, this, throw, try, typeof, var, void, while, with, let, const, await, async},
  keywordstyle=\color{blue}\bfseries,
  ndkeywords={class, export, boolean, throw, implements, import, this},
  ndkeywordstyle=\color{purple}\bfseries,
  identifierstyle=\color{black},
  sensitive=false,
  comment=[l]{//},
  morecomment=[s]{/*}{*/},
  commentstyle=\color{gray}\ttfamily,
  stringstyle=\color{orange}\ttfamily,
  morestring=[b]',
  morestring=[b]"
}
\lstdefinestyle{js}{
  language=js,
  basicstyle=\ttfamily\small,
  keywordstyle=\color{blue}\bfseries,
  commentstyle=\color{gray}\itshape,
  stringstyle=\color{orange},
  showstringspaces=false,
  breaklines=true,
  frame=single,
  captionpos=b
}

\lstdefinelanguage{json}{
  morestring=[b]",
  morestring=[d]',
  stringstyle=\color{orange},
  literate=
   *{:}{{{\color{blue}:}}}{1}
    {,}{{{\color{blue},}}}{1}
    {\{}{{{\color{black}{\{}}}}{1}
    {\}}{{{\color{black}{\}}}}}{1}
    {[}{{{\color{black}[}}}{1}
    {]}{{{\color{black}]}}}{1},
}
\lstdefinestyle{json}{
  language=json,
  basicstyle=\ttfamily\small,
  backgroundcolor=\color{gray!5},
  frame=single,
  showstringspaces=false,
  breaklines=true,
  captionpos=b
}

\lstdefinelanguage{markdown}{
  keywords={\#, \#\#, \#\#\#, \#\#\#\#, -, *, **, _, `, >},
  keywordstyle=\color{teal}\bfseries,
  morecomment=[l]{\%},
  commentstyle=\color{gray},
  morestring=[b]',
  morestring=[b]",
  stringstyle=\color{purple},
  sensitive=false
}
\lstdefinestyle{markdown}{
  language=markdown,
  basicstyle=\ttfamily\small,
  backgroundcolor=\color{gray!10},
  frame=single,
  breaklines=true,
  showstringspaces=false,
  captionpos=b
}

\lstdefinestyle{html}{
  language=HTML,
  basicstyle=\ttfamily\small,
  keywordstyle=\color{blue}\bfseries,
  commentstyle=\color{gray}\itshape,
  stringstyle=\color{orange},
  showstringspaces=false,
  breaklines=true,
  frame=single,
  captionpos=b
}

\lstdefinestyle{bash}{
  language=bash,
  basicstyle=\ttfamily\small,
  keywordstyle=\color{blue}\bfseries,
  commentstyle=\color{gray}\itshape,
  stringstyle=\color{orange},
  showstringspaces=false,
  breaklines=true,
  frame=single,
  captionpos=b
}

\makesavenoteenv{table}
\makesavenoteenv{tabular}

\graphicspath{{images/}}
\titleformat{\section}[block]{\large\bfseries}{\thesection.}{0.5em}{}
\titleformat{\subsection}[block]{\normalsize\bfseries}{\thesubsection}{0.5em}{}
\titleformat{\subsubsection}[block]{\normalsize\itshape}{\thesubsubsection}{0.5em}{}

\title{\textbf{Web-Bench: A LLM Code Benchmark Based on Web Standards and  Frameworks}}
\author[1]{Kai Xu}
\author[1]{YiWei Mao}
\author[1]{XinYi Guan}
\author[1]{ZiLong Feng}
\affil[1]{ByteDance}
\date{}

\begin{document}

\twocolumn[
\maketitle

\begin{abstract}
  The application of large language models (LLMs) in the field of coding is evolving rapidly: from code assistants, to autonomous coding agents, and then to generating complete projects through natural language. Early LLM code benchmarks primarily focused on code generation accuracy, but these benchmarks have gradually become saturated. Benchmark saturation weakens their guiding role for LLMs. For example, HumanEval Pass@1 has reached 99.4\% and MBPP 94.2\%. Among various attempts to address benchmark saturation, approaches based on software engineering have stood out, but the saturation of existing software engineering benchmarks is rapidly increasing. To address this, we propose a new benchmark, Web-Bench \footnotemark \footnotemark, which contains 50 projects, each consisting of 20 tasks with sequential dependencies. The tasks implement project features in sequence, simulating real-world human development workflows. When designing Web-Bench, we aim to cover the foundational elements of Web development: Web Standards and Web Frameworks. Given the scale and complexity of these projects, which were designed by engineers with 5 to 10 years of experience, each presents a significant challenge. On average, a single project takes 4 to 8 hours for a senior engineer to complete. On our given benchmark agent (Web-Agent), SOTA (Claude 3.7 Sonnet) achieves only 25.1\% Pass@1, significantly lower (better) than SWE-Bench's Verified (65.4\%) and Full (33.8\%) scores. Finally, we discuss that in any development field, Standards and Frameworks represent foundational knowledge and efficiency tools, respectively, and LLMs require optimization tailored to them.
\end{abstract}
\vspace{1em}
]
\footnotetext[1]{https://github.com/bytedance/web-bench}
\footnotetext[2]{https://huggingface.co/datasets/bytedance-research/Web-Bench}


\section{Introduction}
The application scope of large language models (LLMs) is constantly expanding and already includes fields such as chatbots, search, and scientific research. Applications in the field of coding are also evolving rapidly, from basic code assistants \cite{GithubCopilot1,GithubCopilot2}, to autonomous coding agents \cite{OpenHands}, and then to attempts to generate complete projects through natural language \cite{WebsiteGen, LLMPlanning}. Although it is still in the early stages of exploration, the potential of LLM coding is already evident despite limited practical results.
\paragraph{}
Early LLM code benchmarks focused on the accuracy of code generation, but these benchmarks have gradually become saturated: HumanEval \cite{HumanEval} SOTA 99.4\% \cite{zhong2024debug} and MBPP \cite{MBPP} SOTA 94.2\% \cite{QualityFlow}. Benchmark saturation weakens their guiding role for LLMs. To solve this problem, many new ideas have emerged, such as increasing programming languages \cite{MultiPL-E,HumanEval-X,HumanEvalPack}, increasing test coverage \cite{CodeContests,EvalPlus}, increasing dataset complexity \cite{DS-1000,CoderEval,BigCodeBench,MLE-bench}, and improving software engineering reproducibility \cite{CodeXGLUE,RepoBench,DevBench,SWE-bench}. We prefer to find a method that can measure the actual complexity of codes. This means that the benchmark should cover the coding process as much as possible and be close to real projects in terms of logical complexity. Among many ideas, software engineering is undoubtedly the closest to this goal.
\paragraph{}
Existing software engineering benchmarks, such as SWE-Bench \cite{SWE-bench} and RepoBench \cite{RepoBench}, both use independent tasks to organize datasets. Although the data in these benchmarks may come from real projects, the tasks themselves lack coherence and differ significantly from real-world development workflows. This paper proposes a new benchmark, Web-Bench, which contains 50 projects, each consisting of 20 tasks with sequential dependencies. The tasks implement project features in sequence, simulating real-world human development workflows. When designing Web-Bench, we aim to cover the foundational elements of Web development: Web Standards and Web Frameworks. Web Standards are basic knowledge of Web development and cover CSS Selector, CSS Layout, DOM, SVG, WebGL, ECMAScript \cite{JavaScript}, TypeScript \cite{Typescript}, etc. Web Frameworks provide higher-level abstraction capability and improve development efficiency, covering key aspects of Web development such as UI Framework \cite{Reactjs}, State Management \cite{State}, CSS Framework, Fullstack Framework, ORM, Build, etc. 
\paragraph{}
Given the scale and complexity of these projects, which were designed by engineers with 5-10 years of experience, each presents a significant challenge. On average, a single project takes 4–8 hours for a senior engineer to complete. On our given benchmark agent (Web-Agent), SOTA (Claude 3.7 Sonnet) achieves only 25.1\% Pass@1, significantly lower than SWE-Bench's 65.4\% (Verified) \cite{SWE-agent} and 33.8\% (Full). A year ago (2024.4),  SWE-Bench (Verified) SOTA was only 25.6\% which has increased rapidly. Table \ref{tab:sotas} below compares the SOTA of more benchmarks.

\section{Benchmark Design}
\subsection{Benchmark Construction}
\begin{figure}[htbp!]
  \centering
  \includegraphics[width=1\linewidth]{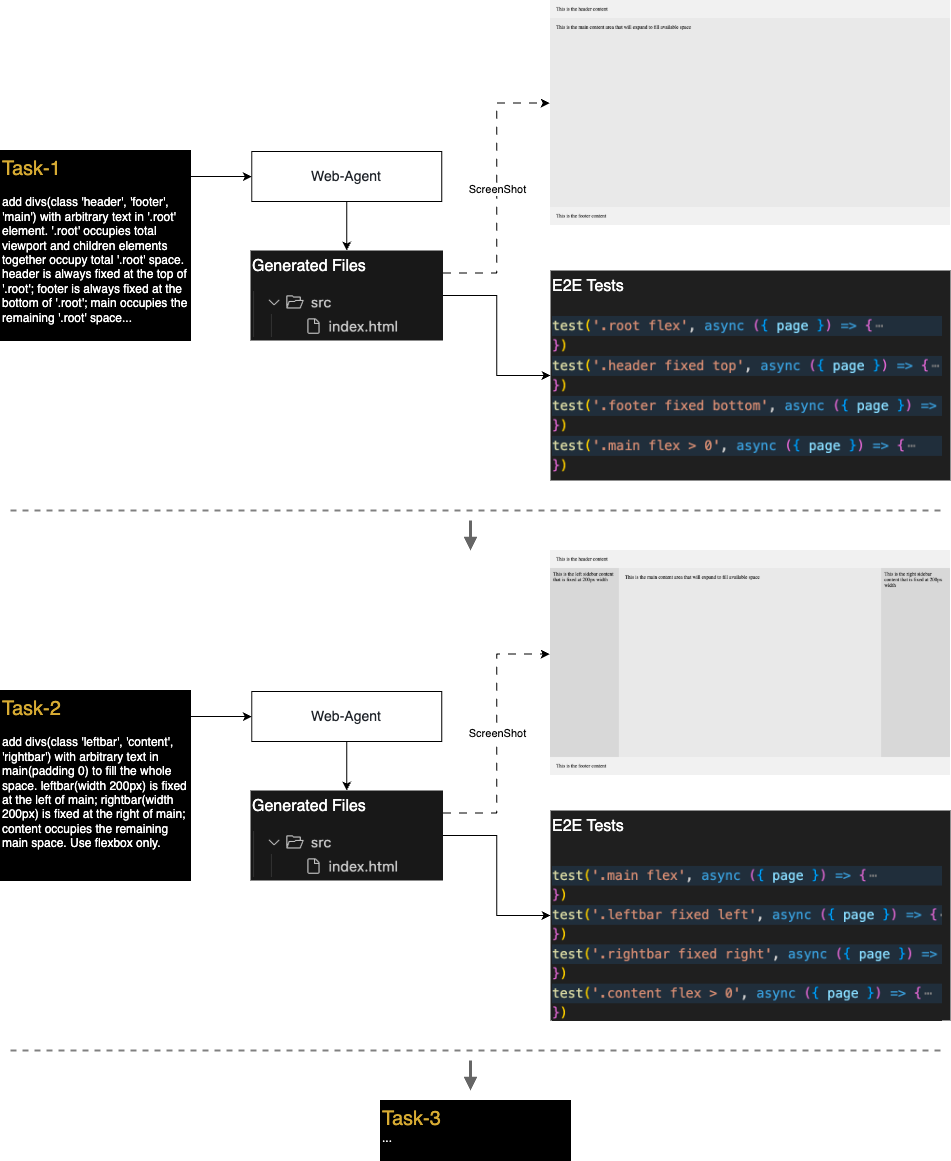}
  \caption{Task-2 depends on the execution result ('main' element) of Task-1}
  \label{fig:benchmark-construction}
\end{figure}

The Web-Bench dataset contains 50 projects, each consisting of 20 tasks with sequential dependencies. See Figure \ref{fig:benchmark-construction}.
\begin{enumerate}
  \item Dataset: a collection of projects. Due to the evaluation cost, we plan to provide a Lite version in the future.
  \item Project: a single item in the dataset.
  \item Task: including a description and several end-to-end (E2E) testcases. On average, there are 3.6 cases per task and 72.4 cases per project, as shown in Table \ref{tab:project-attributes}.
\end{enumerate}

\begin{table}[htbp!]
  \centering
  \caption{Average and maximum attributes of across 50 projects}
  \begin{tabular}{llcc}
  \toprule
  Project & Attribute & Average & Maximum \\
  \midrule
  \multirow{2}{*}{Solution} & Files & 23.4 & 76 \\
    & Lines & 1947.9 & 6180 \\
  \multirow{2}{*}{Testcases} & PerTask & 3.6 & 5 \\
    & Total & 72.4 & 99 \\
  \bottomrule 
  \end{tabular}
  \label{tab:project-attributes}
\end{table}

\begin{figure}[htbp!]
  \centering
  \includegraphics[width=0.8\linewidth]{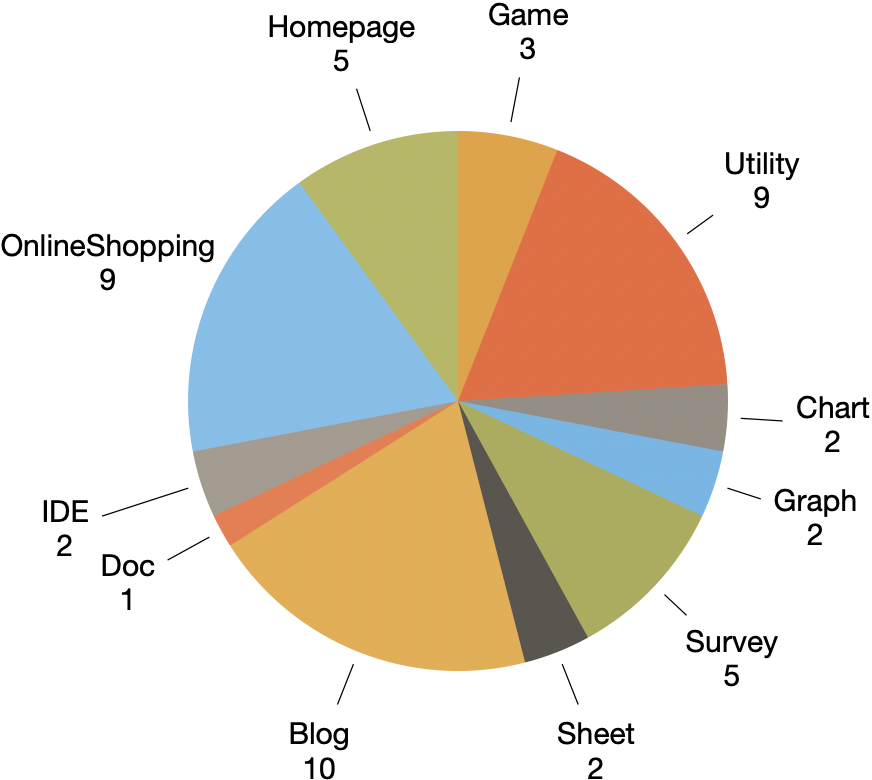}
  \caption{Distribution of projects across scenarios}
  \label{fig:project-scenarios}
\end{figure}

\paragraph{}
During the design process of the project, different business scenarios \ref{fig:project-scenarios} were selected: Games, Utilities, Charts, Drawing Boards, Surveys, Data Tables, Blog Sites, Document Sites, etc., so that the final Benchmark is closer to real-world software engineering. 

\paragraph{}
Web-Bench covers the core Web standards and mainstream development frameworks \ref{fig:project-domains} \ref{fig:standards-frameworks}  that Web application development relies on. see Appendix \ref{appendix:a} for more details.
\begin{figure}[htbp!]
  \centering
  \includegraphics[width=0.5\linewidth]{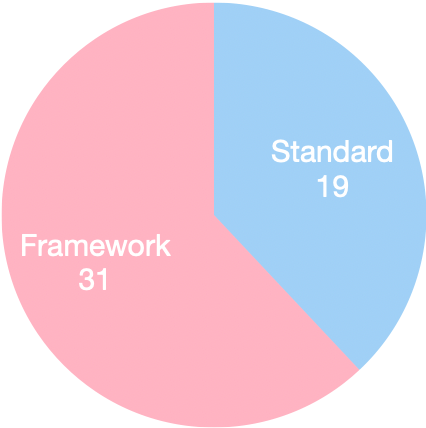}
  \caption{Distribution of projects across domains}
  \label{fig:project-domains}
\end{figure}
\begin{figure}[htbp!]
  \centering
  \includegraphics[width=1\linewidth]{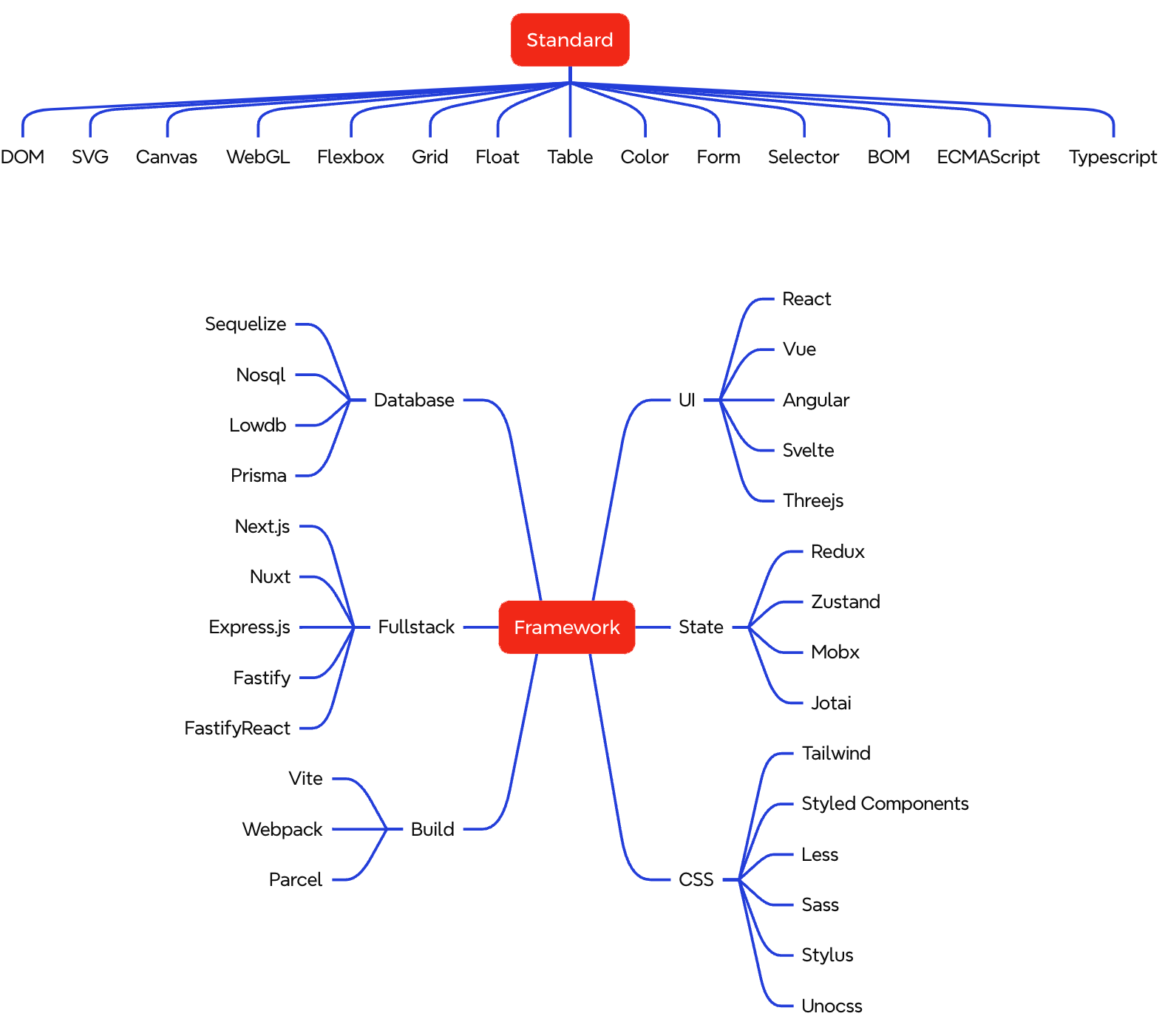}
  \caption{All Standards and Frameworks in Web-Bench}
  \label{fig:standards-frameworks}
\end{figure}

\subsection{Standard and Framework}
\subsubsection{Standard}
Web Standards are the basic knowledge of Web development, especially those that are frequently used, offer extensive functionality, and enable more complex features. We have selected the core standards that Web applications rely on, covering HTML, CSS and JavaScript.

\paragraph{}
Partial core standards:
\begin{itemize}
  \item ECMAScript: the JavaScript language standard that ensures the interoperability of web pages across different web browsers.
  \item TypeScript: A strongly typed programming language that builds on JavaScript. TypeScript code converts to JavaScript, which runs anywhere JavaScript runs. Most modern web applications are now built with  TypeScript.
  \item CSS Selector: Patterns that match against elements in a tree and are one of the core technologies used to select nodes in a document.
  \item Flexbox: A layout model that arranges items in one dimension. The children of a flex container can be laid out in any direction, and can "flex" their sizes, either growing to fill unused space or shrinking to avoid overflowing the parent. It is widely used in most web applications.
  \item Grid: A two-dimensional grid-based layout system. The children of a grid container can be positioned into arbitrary slots in a predefined flexible or fixed-size layout grid. It is more powerful than flexbox and also widely used in web applications.
  \item DOM (Document Object Model): defines a platform neutral model for events and node trees. It is the main technology used to manipulate web page elements.
  \item SVG (Scalable Vector Graphics): An XML-based language for describing two-dimensional graphics. SVG drawings can be interactive and dynamic. Animations can be defined and triggered declaratively.
  \item Canvas: Used for drawing graphics via JavaScript and the HTML <canvas> element. It can be used for animation, game graphics, data visualization, photo manipulation, and real-time video processing.
  \item WebGL (Web Graphic Language): An immediate mode 3D rendering API designed for the web.  It is derived from OpenGL® ES 3.0, and provides similar rendering functionality, but in an HTML context.
\end{itemize}

\subsubsection{Framework}
In the early stages of Web development, there were not many development frameworks, and various Web applications could be developed using native technologies. However, with the rise of Web 2.0, both the technical complexity and application scale of Web applications have increased dramatically, and the development efficiency based solely on native Web technology has been challenged. Development frameworks have brought greater abstraction capabilities, improved development efficiency, and enhanced collaboration to Web development, covering its main aspects:

\begin{itemize}
  \item UI Framework: Usually provides a component-based development model and supporting development tools to improve the efficiency of single-unit development and collaboration, such as React, Vue, Angular, Threejs, etc.
  \item State Management: UI state management is one of the core tasks in Web application development. Excellent frameworks will simplify the state model and improve development and maintenance efficiency, such as Redux, Zustand, Jotai, Mobx, etc. 
  \item CSS Framework: Improves CSS maintainability and development efficiency by enhancing basic capabilities such as variables, modules, functions, or providing CSS utilities. Examples include SASS, LESS, Stylus, Tailwind, etc.
  \item Build: Web applications eventually run in browsers, and during the development phase, JS code and various resources (CSS, images, icons etc.) are usually split into many modules (or files), which require modular packaging using build tools, as well as optimization such as merging and compression. At the same time, build tools are also required to unify the development experience — handling JS syntax conversion, CSS preprocessing, and integration of various tools (including linting, formatting, testing frameworks, etc.) — such as Webpack, Vite, Parcel, etc.
  \item Fullstack Framework: Fullstack usually refers to the ability of developers to handle both front-end (such as UI) and back-end (such as data and logic processing) development. In this context, it refers to frameworks that support both front-end and back-end development, such as Next.js, Nuxt, Express.js, etc.
  \item Database Framework Data management is a relatively independent and core part of back-end logic. Data can be managed directly via SQL using tools like SQLite, or through object-relational mapping frameworks such as Prisma and Sequelize.
\end{itemize}

\subsubsection{Calibrate Project}
Use Web-Bench evaluation tool to Calibrate a project. More details can be found in Appendix \ref{appendix:d}.
That is, check for design issues in the project:
\begin{enumerate}
  \item Task issues, such as ambiguous descriptions, vulnerabilities, or obvious conflicts with testcases.
  \item Testcase issues, such as not rigorous enough or too strict.
\end{enumerate}

\section{Benchmark Evaluation}
\subsection{Project Structure}
\label{sec:project-structure}
\begin{lstlisting}[style=bash]
src/               # final solution codes
  index.html
src-init/          # initial solution codes
  index.html 
test/
  init.spec.js     # cases for initial codes
  task-1.spec.js   # cases for task-1
  ...
  task-20.spec.js  # cases for task-20
tasks.jsonl        # support json and yaml
\end{lstlisting}

\subsection{Evaluator Workflow}
\label{sec:evaluator-workflow}

\begin{figure}[htbp!]
  \centering
  \includegraphics[width=1\linewidth]{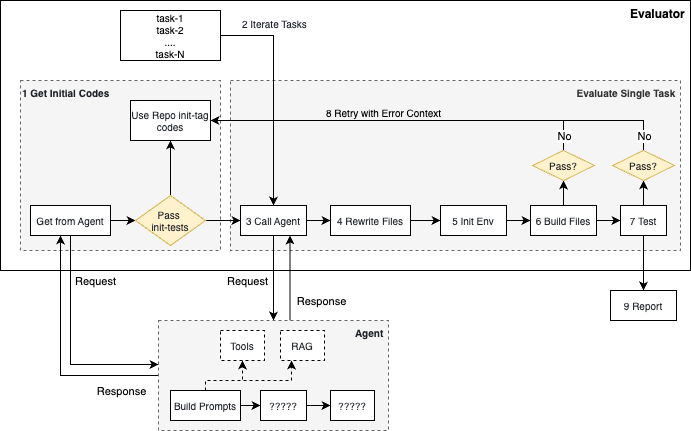}
  \caption{Evaluator Workflow}
  \label{fig:evaluator-workflow}
\end{figure}
\begin{enumerate}
  \item  Get Initial Codes: optional step, get initial codes from agent.
  \item  Iterate Tasks: from task-1 to task-20.
  \item  Call Agent: more details about agent can be found in Section \nameref{sec:web-agent-workflow}.
  \item  Rewrite files: update files or create new files.
  \item  Init Env: optional step,init files running environment.
  \item  Build Files: optional step, check files errors, such as  reference errors.
  \item  Test: end-to-end (E2E) test with Playwright.
  \item  Retry: with error context from Build or Test.
  \begin{enumerate}
    \item First attempt, if failed, implement task-n with task description and error context, goto step-3.
    \item Second attempt, if failed, evaluation is done.
  \end{enumerate}
  \item  Report.
\end{enumerate}
More details can be found in Appendix \ref{appendix:c}.

\subsection{Web-Agent Workflow}
\label{sec:web-agent-workflow}
Web-Agent is an instance of the agent described in Section 3.2, Step 3.  Web-Agent is the interaction module between Web-Bench and LLM. Its main responsibilities are to build prompts, request LLM API, and extract files from LLM response.
\begin{figure}[htbp!]
  \centering
  \includegraphics[width=0.8\linewidth]{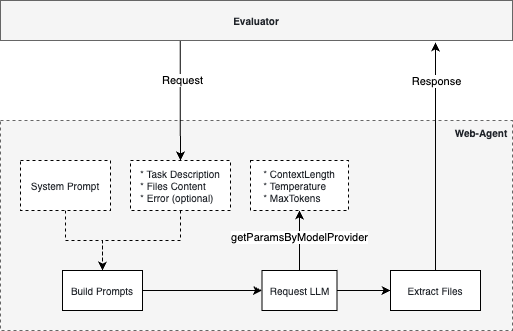}
  \caption{Web-Agent Workflow}
  \label{fig:web-agent-workflow}
\end{figure}
\begin{enumerate}
  \item Build Prompts: Using system prompt (SP), task description, files, and error messages. If the combined input exceeds the context length, it will be truncated.
  \item Request LLM: As shown in Table \ref{tab:model-parameters}. Requests the LLM's OpenAPI based on the selected model and returns the response. In this step, parameters such as temperature, maxToken, contextLength are obtained from the model provider.
  \item Extract Files: Parses the LLM's response to extract the generated files.
\end{enumerate}
More details can be found in Appendix \ref{appendix:b}.

\noindent
\begin{table*}[htbp]
  \centering
  \caption{Model Parameters}
  \begin{tabular}{lccc}
  \toprule
    Model Vendor & contextLength & maxToken & temperature \\
  \midrule
  anthropic & 50,000 & 8,192 & 1 \\
  qwen & 10,000 & 4,096 & default \\
  deepseek & 10,000 & 8,192 & 0 \\
  openai & 10,000 & 8,192 & 0.4 \\
  doubao & 10,000 & 4,096 & default \\
  openrouter & 10,000 & 8,192 & 0.4 \\
  \bottomrule 
  \end{tabular}
  \label{tab:model-parameters}
\end{table*}

\subsection{Metrics}
\label{sec:metrics}

\begin{equation}
\textbf{pass@1} = \left(\frac{\textbf{pass\_tasks@1}}{\textbf{tasks}}\right) \times 100\%
\end{equation}

\begin{itemize}
  \item Percentage of tasks passing all E2E tests on the first attempt.
  \item pass\_tasks@1: number of tasks passed on first attempt before any failure.
  \item tasks: total number of tasks in 'tasks.jsonl'.
\end{itemize}

\begin{equation}
\textbf{pass@2} = \left(\frac{\textbf{pass\_tasks@2}}{\textbf{tasks}}\right) \times 100\%
\end{equation}

\begin{itemize}
  \item Percentage of tasks passing all E2E tests on the second attempt. pass@2 (retry with error context) is closer to the behavior of human engineers.
  \item pass\_tasks@2: total number of passed tasks including retries.
\end{itemize}

Example:
\begin{itemize} 
  \item tasks: 20
  \item pass\_tasks@1: 5, the first failed evaluation happened at task-6
  \item pass\_tasks@2: 15, total passed tasks is 15 (task-16's second attempt failed)
  \item pass@1: $5 / 20 \times 100\% = 25\%$
  \item pass@2: $15 / 20 \times 100\% = 75\%$
\end{itemize}

\section{Results}

\subsection{Evaluation Results}
\begin{figure}[htbp!]
  \centering
  \includegraphics[width=1\linewidth]{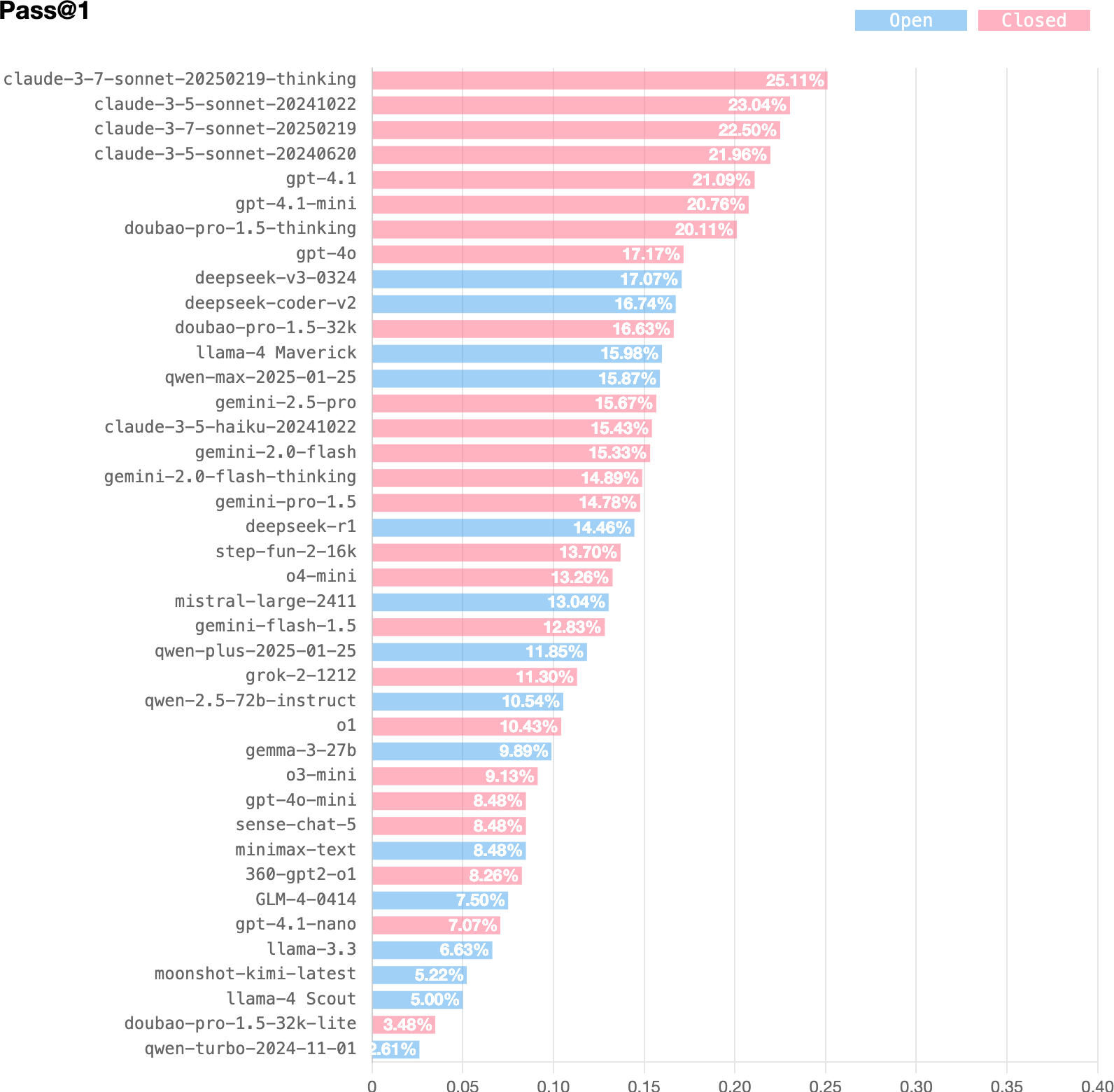}
  \caption{Pass@1 across all models.}
  \label{fig:pass-1}
\end{figure}

\begin{figure}[htbp!]
  \centering
  \includegraphics[width=1\linewidth]{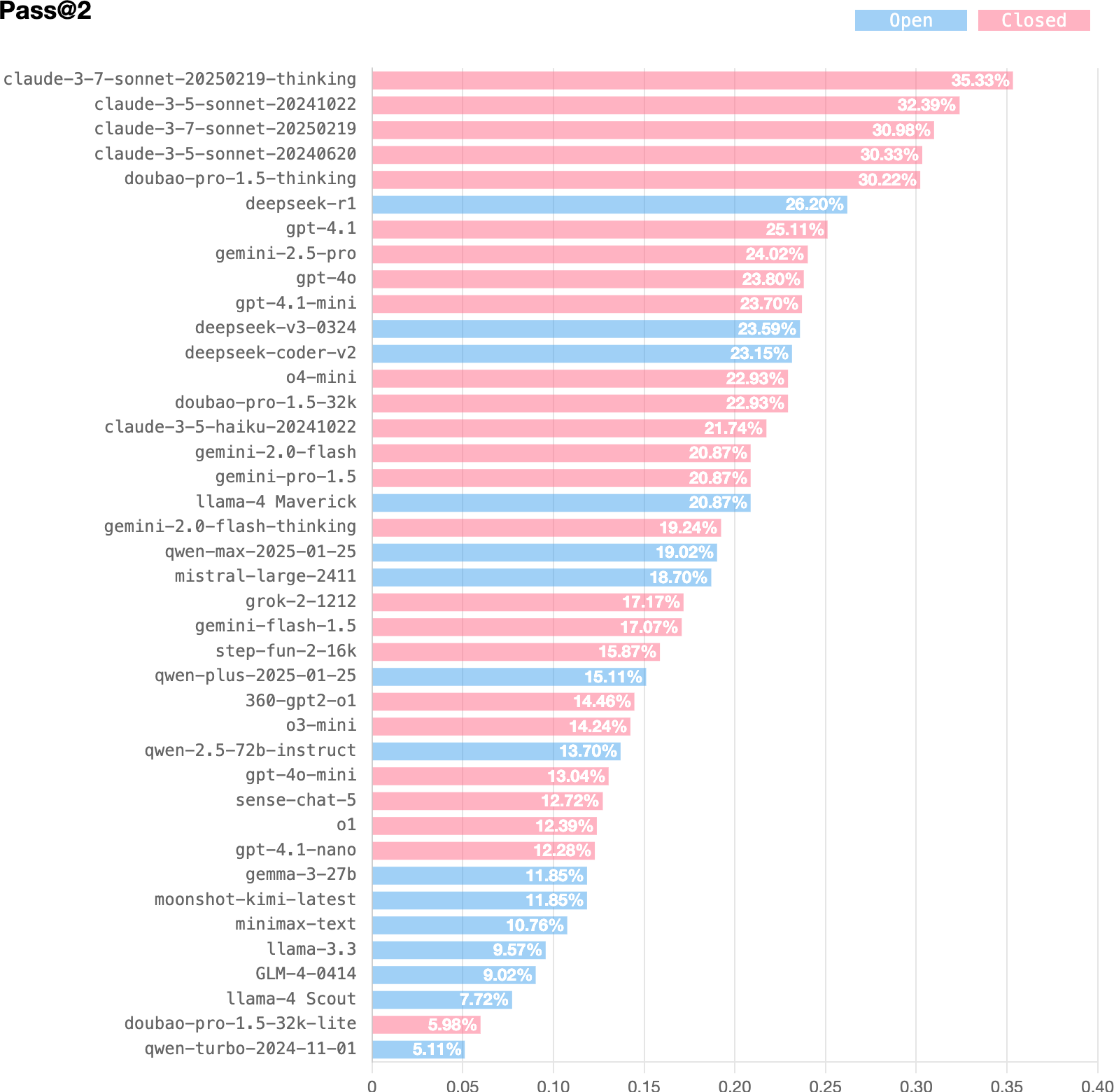}
  \caption{Pass@2 across all models.}
  \label{fig:pass-2}
\end{figure}

\begin{figure}[htbp!]
  \centering
  \includegraphics[width=1\linewidth]{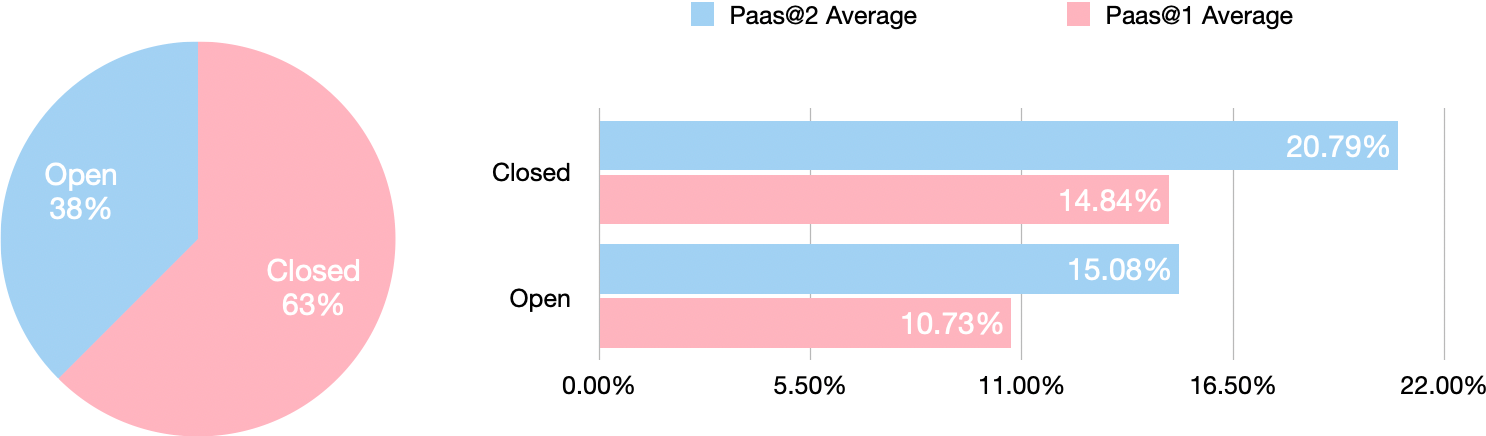}
  \caption{Average Pass@1 and Pass@2 by model type (Open vs. Closed)}
  \label{fig:open-closed}
\end{figure}

As shown in Figure \ref{fig:pass-1}, 'claude-3-7-sonnet-20250219-thinking' achieves the highest Pass@1 and Pass@2 (using Best-of-5 sampling, as used throughout). From the Pass@1 distribution, Claude-series models exhibit the highest performance, followed by GPT, Doubao, DeepSeek, LLAMA, Gemini, Qwen, and others. From the Pass@2 distribution, Claude-series models exhibit the highest performance, followed by Doubao, DeepSeek, GPT, Gemini, LLAMA, Mistral, and others.
\paragraph{}
As shown in Figure \ref{fig:pass-1} and \ref{fig:pass-2}, regardless of Pass@1 or Pass@2, closed-source models (shown in red) also appear to have an advantage. As shown in Figure \ref{fig:open-closed}, the average Pass@2 of closed models is 20.79\%, while that of open models is 14.84\%. The average Pass@1 of closed models is 15.08\%, while that of open models is 10.73\%. 
\paragraph{}
More details can be found in Appendix \ref{appendix:e}.

\begin{figure}[htbp]
  \centering
  \includegraphics[width=1\linewidth]{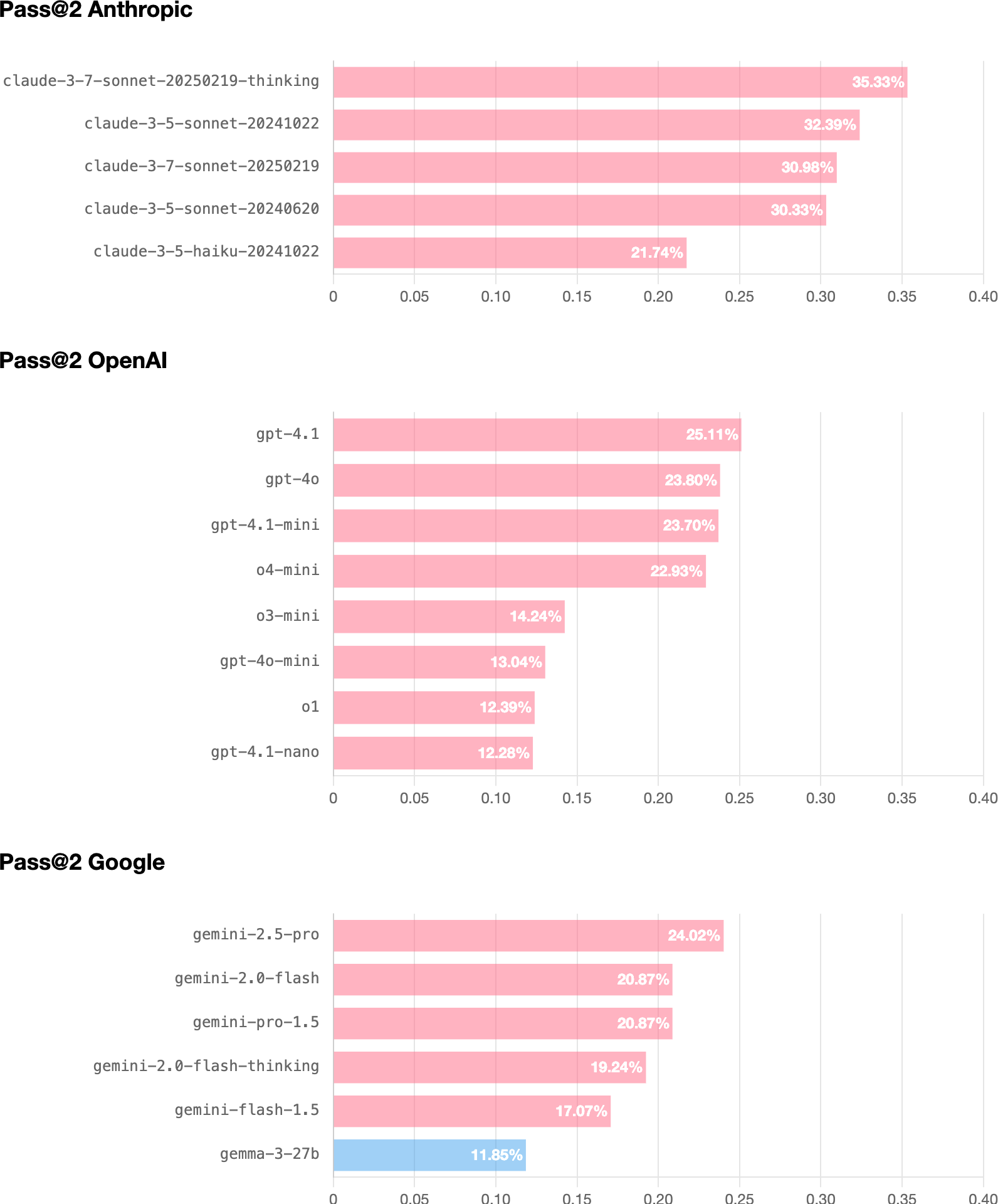}
  \caption{Pass@2 across LLMs within the same model series - Part 1}
  \label{fig:pass-2-series}
\end{figure}

\begin{figure}[htbp]
  \centering
  \includegraphics[width=1\linewidth]{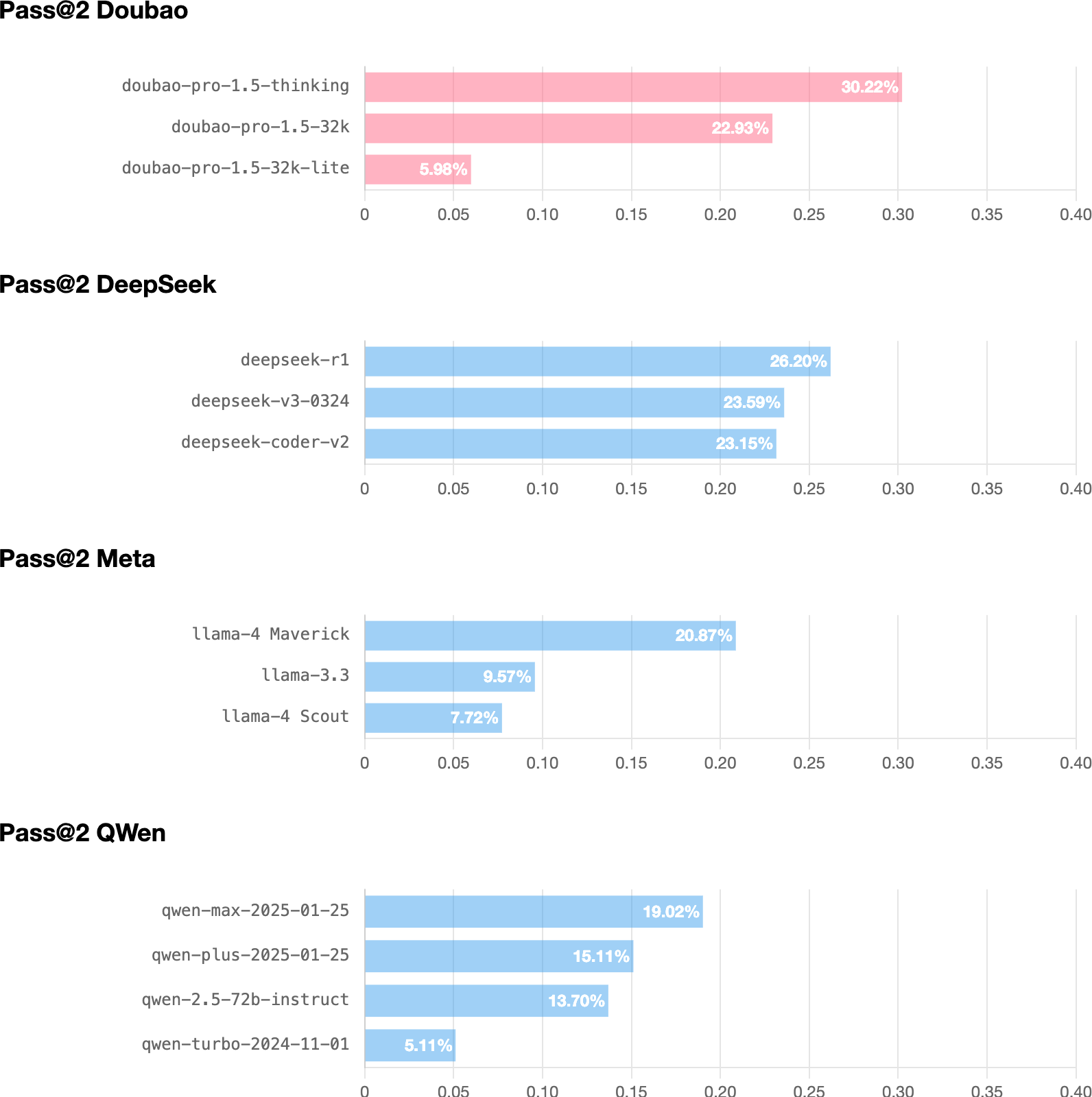}
  \caption{Pass@2 across LLMs within the same model series - Part 2}
  \label{fig:pass-2-series-2}
\end{figure}

\paragraph{}
As shown in Figure \ref{fig:pass-2-series} and \ref{fig:pass-2-series-2}, the Pass@2 distribution across LLM series indicates that newer or larger-parameter models generally perform better in Web-Bench, which aligns with the commonly observed scaling law.

\begin{figure}[htbp!]
  \centering
  \includegraphics[width=1\linewidth]{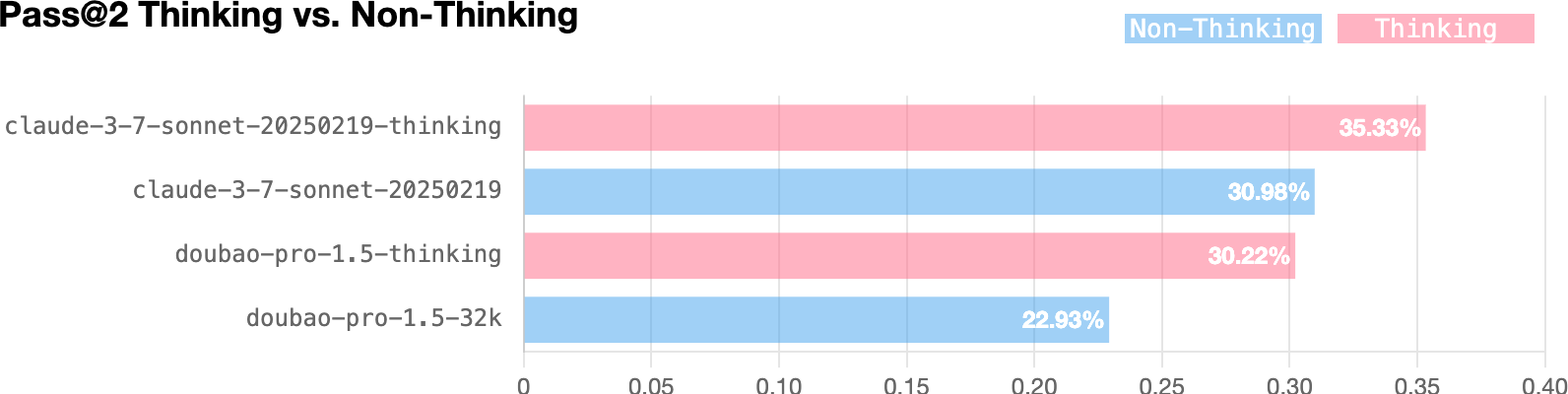}
  \caption{Pass@2 across LLMs with and without thinking}
  \label{fig:pass-2-thinking}
\end{figure}

\paragraph{}
As shown in Figure \ref{fig:pass-2-thinking}, LLMs of the same specification perform better on Pass@2 when the thinking mode is enabled.

\subsection{Compare SOTAs}
\begin{figure}[htbp!]
  \centering
  \includegraphics[width=1\linewidth]{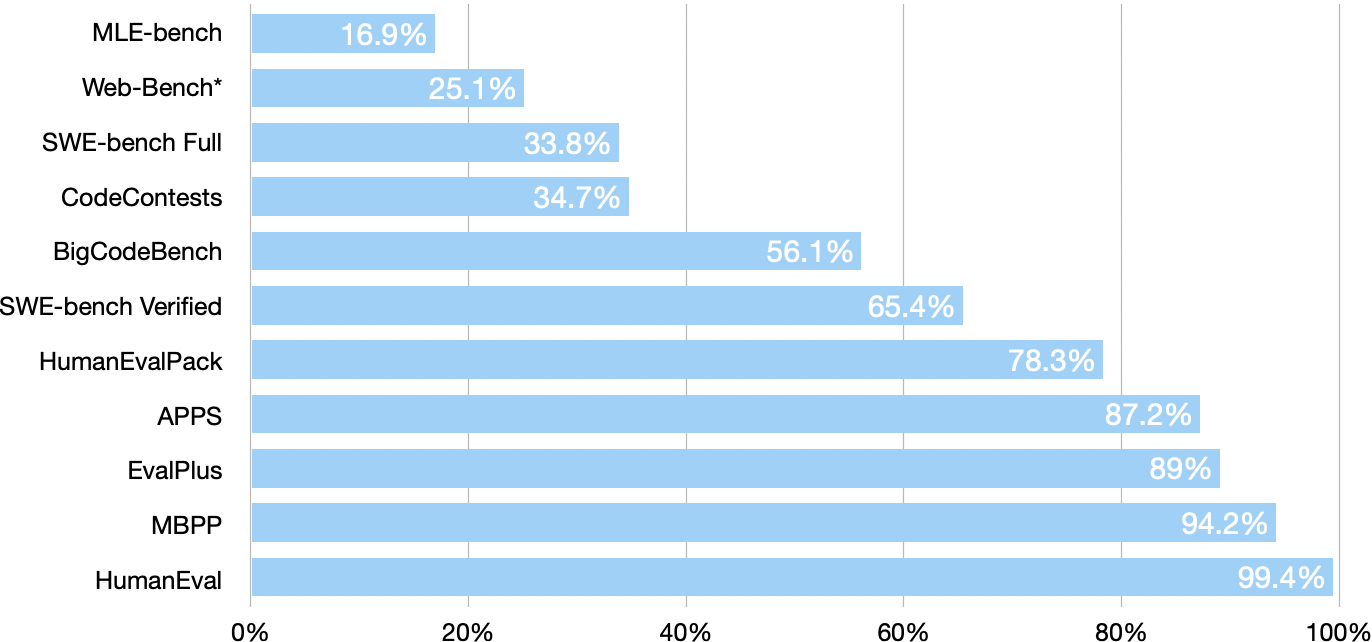}
  \caption{Benchmark SOTA Pass@1 as of April 2025}
  \label{fig:sotas}
\end{figure}

As shown in Table \ref{tab:sotas} and Figure \ref{fig:sotas}, HumanEval and MBPP have approached saturation. APPS and EvalPlus are approaching saturation. The SOTA for Web-Bench is 25.1\%, which is lower (better) than that of the SWE-bench Full and Verified sets.

\begin{table*}[htbp]
  \centering
  \caption{Benchmark SOTA Pass@1 as of April 2025}
  {\scriptsize
  \begin{tabular}{lccl}
  \toprule
  Benchmark & Pass@1 & Date \footnotemark & SOTA \\
  \midrule
  MLE-bench & 16.9\% & 2024.10 & AIDE o1-preview \cite{MLE-bench} \\
  Web-Bench$^*$ & 25.1\% & 2025.4 & Web-Agent (Claude 3.7 Sonnet) \\
  SWE-bench Full & 33.8\% & 2025.2 & SWE-agent (Claude 3.7 Sonnet) \cite{SWE-agent}\\
  CodeContests & 34.7\% & 2024 & LPW (GPT-4o) \cite{CodeContests} \\
  BigCodeBench & 56.1\% & 2025.2 & DeepSeek-V3 \\
  SWE-bench Verified & 65.4\% & 2025.3 & Augment Agent v0 \\
  HumanEvalPack & 78.3\% & 2024.2 & GPT-4 \\
  APPS & 87.2\% & 2025.1 & LPW (GPT-4o) \cite{PlanningDriven} \\
  EvalPlus & 89.0\% & 2024.9 & O1 Preview \\
  MBPP & 94.2\% & 2025.3 & QualityFlow (Claude 3.5 Sonnet) \cite{QualityFlow} \\
  HumanEval & 99.4\% & 2024.6 & LLMDebugger (OpenAI o1) \cite{zhong2024debug}  \\
  \bottomrule 
  \end{tabular}
  }
  \label{tab:sotas}
\end{table*}
\footnotetext{Paper / Agent / LLM Publication Date}

\section{Related Work} 
Early LLM code benchmarks focused on the accuracy of single-function code generation. We review several widely cited benchmarks:
\begin{itemize}
  \item HumanEval \cite{HumanEval}: 164 handwritten Python coding questions are used to evaluate functional correctness. Each question contains a function signature, docstring, and function body, and has an average of 7.7 test cases.
  \item MBPP \cite{MBPP}: About 1,000 entry-level crowd-sourced Python coding questions covering coding basics and standard library functions, with each question having an average of 3 test cases.
  \item APPS \cite{APPS}: 10,000 Python coding questions—5,000 for training and 5,000 for testing—sourced from multiple coding competition websites, covering simple problems, technical interview questions, and algorithm competition-level challenges. Each question has an average of 13 test cases.
\end{itemize}
\paragraph{}
From Table \ref{tab:sotas}, we can see that HumanEval is saturated, and MBPP is close to saturation. APPS performs slightly better due to its larger dataset size, but it is also approaching saturation. To address code benchmark saturation, many new ideas have emerged.

\subsection{Increase programming languages}
Most of the problems in early code benchmarks used Python as the programming language. In addition to Python, mainstream languages in real-world development include C++, JavaScript, Java, Go, etc., making it natural to consider expanding language coverage. 
\begin{itemize}
  \item MultiPL-E \cite{MultiPL-E} translates HumanEval and MBPP to 18 other programming languages.
  \item HumanEval-X \cite{HumanEval-X} extends the HumanEval benchmark to Python, C++, Java, JavaScript, and Go via human translations
  \item HumanEvalPack \cite{HumanEvalPack} extends the HumanEval benchmark to cover three coding tasks—code repair, code interpretation, and code generation—and six languages (Python, JavaScript, Java, Go, C++, Rust) via human translations. As shown in Table \ref{tab:sotas}, the SOTA in 2024 has reached 78\%.
\end{itemize}

\subsection{Increase test coverage}
This stems from a simple idea: improving test coverage and increasing the presence of edge cases is expected to reduce the pass rate.
\begin{itemize}
  \item CodeContests \cite{CodeContests} is a competitive coding dataset for machine learning. It automatically generates additional test cases (mutations), such as bit flipping of input strings or integer addition and subtraction. It uses multiple correct solutions to verify the new tests and ensure consistent output, and filters out problems with insufficient or overly simple test cases. 
  \item EvalPlus \cite{EvalPlus} detects the robustness of generated code by strengthening the scope and number of HumanEval and MBPP tests. For the same LLM, Pass@1 can generally be reduced by 5–10\%. However, as shown in Table \ref{tab:sotas}, the latest SOTA has reached 89\%, gradually approaching saturation.
\end{itemize}

\subsection{Increasing dataset complexity}
The problems in the early code benchmarks contain many basic algorithms, which are no longer challenging for LLMs; increasing the complexity of a single problem is also an idea. 
\begin{itemize}
  \item DS-1000 \cite{DS-1000} is a code generation benchmark containing 1,000 data science problems, covering seven Python libraries such as NumPy and Pandas. The problems come from Stackoverflow and have been lightly edited. 
  \item CoderEval \cite{CoderEval} contains 230 Python and 230 Java code generation problems carefully selected from popular open source projects. It supports code generation problems with six different levels of context dependency. It supports problems with six levels of context dependency, and performance is significantly better when generating independent functions compared to non-independent ones.
  \item BigCodeBench \cite{BigCodeBench} combines tools from 139 libraries across 7 domains to support 1,140 fine-grained tasks. It integrates multiple existing benchmarks and enables the implementation of functions such as data analysis and web development by calling multiple tools.
  \item MLE-bench \cite{MLE-bench} is a benchmark for measuring the performance of AI agents in machine learning engineering tasks. 75 ML engineering competitions were selected from Kaggle to evaluate real-world skills such as model training, dataset preparation, and running experiments for machine learning workflows. As shown in Table \ref{tab:sotas}, the current SOTA is only 16.9\%. As the coding capabilities of LLMs continue to improve, such professionally oriented benchmarks are expected to become more prevalent.
\end{itemize}

\subsection{Improving software engineering reproducibility}
Some benchmarks simulate the software engineering process from different perspectives:
\begin{itemize}
  \item CodeXGLUE \cite{CodeXGLUE} includes 14 datasets for 10 diverse programming language tasks including code-code, text-code, code-text and text-text scenarios. Provides three baseline models to support these tasks, including a BERT-style pre-trained model for understanding tasks, a GPT-style pre-trained model for code completion and generation, and an encoder-decoder framework that supports sequence-to-sequence generation.
  \item RepoBench \cite{RepoBench} is a new benchmark containing three subtasks (RepoBench-R retrieval, RepoBench-C completion, RepoBench-P pipeline), supporting Python and Java, and covering the evaluation of automatic code completion systems at the code repository level.
  \item DevBench \cite{DevBench} is based on the waterfall model. It explores the performance of the evaluated model across five stages: software design, environment configuration, implementation, acceptance testing, and unit testing.
  \item SWE-bench \cite{SWE-bench} focuses on solving real and complex software engineering problems. 2,294 tasks come from 12 popular Python libraries on GitHub. The filtering rules are as follows: approximately 90,000 Pull Requests were collected, each associated with a code base. A GitHub issue was resolved, the test files in the repository were modified, at least one test case was included, and the test status changed from failed to passed. SWE-bench offers substantial scale and real-world software engineering complexity. In the past two years, it has gradually become a highly cited code benchmark. As shown in Table \ref{tab:sotas}, after two years, the SOTA—especially on the Full set—remains far from saturation.
\end{itemize}

\section{Discussion}
\subsection{Why Software Engineering? }
Based on the discussion in Chapter 5 and the SOTA data presented in Table \ref{tab:sotas}:

\paragraph{}
Increasing the types of programming languages cannot solve the benchmark saturation problem. This is intuitive. Real-world programming languages have their own characteristics and strengths, but are similar in expressiveness. In fact, LLM pass rates across different language versions of the same benchmark do not vary significantly \cite{HumanEvalPack}.
\paragraph{}
Increasing test coverage seems to have some effect, but EvalPlus \cite{EvalPlus} also approached saturation within two years of its publication. Although CodeContests \cite{CodeContests} performs better, its unsaturated SOTA is likely due to the high difficulty of its competition-level question set. Similarly, in real-world scenarios, project complexity arises from factors such as lines of code, code logic complexity, and the structure of the module/component dependency tree. Even 100\% test coverage cannot change the complexity of the project itself.
\paragraph{}
Increasing the complexity of the dataset—whether by increasing the complexity of individual problems \cite{CoderEval,MLE-bench} or by expanding the scale of the benchmark \cite{BigCodeBench}—appears effective based on SOTA results. However, the upper limit of this approach often consists of uncommon and complex algorithmic problems, which gradually deviate from real-world coding. This may lead LLM coding to evolve in a less practical direction.
\paragraph{}
Simulating the software engineering process has been theoretically demonstrated as feasible by CodeXGLUE, RepoBench, and DevBench. These benchmarks have also been preliminarily validated on earlier LLMs, paving the way for larger-scale software engineering benchmarks such as SWE-bench and Web-Bench. Benchmarks in the software engineering direction are grounded in real-world coding behaviors.

\paragraph{}
Web-Bench is a new software engineering benchmark and fills the gap: Sequential task evaluation for fullstack web development based on Web standards and frameworks. \ref{tab:benchmarks-evolution}

\begin{table*}[htbp]
  \centering
  \caption{Benchmarks Evolution}
  \begin{tabular}{lllll}
  \toprule
  Benchmark & Task Type & Dependency & Tech Stack & Evaluation Type \\
  \midrule
  HumanEval & Single Function & No & Python & Function Test \\
  SWE-bench & Issue Fixing / Feature & Low & Python & Unit Test \\
  Web-Bench & Project & Strong & Web Fullstack & E2E Test \\
  \bottomrule 
  \end{tabular}
  \label{tab:benchmarks-evolution}
\end{table*}

\subsection{LLMs require optimization tailored to Standards and Frameworks}
Referring to the previous section, we note that software engineering provides a promising solution to benchmark saturation. Furthermore, if the problems contained in the benchmark (or in the projects presented in this work) accurately reflect real-world code, they can help guide LLMs to evolve in a practical direction. A fundamental question arises: what is the basis of real-world coding? Although this article primarily discusses web application development, the development process in any field \cite{SoftwareArchitecture} is fundamentally built upon standards (also referred to as protocols, specifications, or recommendations) and frameworks (which may include libraries or third-party packages). Standards and Frameworks represent the basic knowledge and efficiency tools in a coding field, respectively. From this perspective, they are universally applicable. So we have an intuition: LLM needs optimization tailored to Standards and Frameworks. 
\paragraph{}
We are also conducting further in-depth research in this direction, which may serve as a potential breakthrough in the next stage: 
\begin{enumerate}
  \item Standards-aware Training:
  \begin{itemize}
    \item Inject W3C/WHATWG documentation into pretraining data.
    \item Finetune with AST based syntax constraint checks.
  \end{itemize}
  \item Framework-specific Adaptation:
  \begin{itemize}
    \item Add framework rule checkers to decoding loops.
    \item Develop plugin architectures for framework updates (avoiding retraining).
  \end{itemize}
\end{enumerate}

\section{Limitation}
Web 2.0 has brought about the rapid development of the Web development ecosystem. Dating back to 2005, nearly two decades have passed. Although the current version of Web-Bench covers the core Standards and Frameworks, many others remain uncovered due to workload constraints. In the future, we will strive to cover more standards and frameworks by innovating project design and development tools. 
\paragraph{}
This paper mainly studies the accuracy of code generation for LLMs. We have also received feedback from users on multiple occasions expressing interest in evaluating the aesthetics and code quality of the generated pages. Since the evaluation of aesthetics and code quality can largely reuse the existing benchmark infrastructure, Web-Bench already includes a built-in screenshot function. In the future, it can be easily extended to support additional metrics.

\section*{Acknowledgements}
We thank YuBei Li, Quan Ding, Feng Gu, YuKun Wang, ChaoMo Li, Kai Shen, ShuLin Xin, LinJian Cao, and KaiGe Yu for helpful discussions and feedback on drafts of this work. We are also grateful to Seed Team for their work on software and hardware infrastructure used by this benchmark. We thank the Continue project for inspiring our web-agent, as well as all the open-source projects used in this work, including Node, VSCode, RushStack, Vite, Playwright, React, Next.js, and more. Finally, we would like to thank all the organizations and individuals who contribute to web standards. Their work forms the foundation of the entire web development ecosystem, including but not limited to W3C, WHATWG, and TC39.

\medskip
\bibliographystyle{unsrt} 
\bibliography{references}

\begin{table*}[ht]
  \centering
  \caption{Projects for Web Standards}
  {\small
  \begin{tabular}{l l p{10cm} l}
    \toprule 
    Project & Domain & Description & Scenario \\ 
    \midrule 
    Table & CSS & The element represents tabular data that is information presented in a two-dimensional table comprised of rows and columns of cells containing data. It was used for layout in the early days of the Web, but now it is only recommended for displaying data. & Sheet \\ 
    Flex & CSS & The layout of items in one dimension. The children of a flex container can be laid out in any direction, and can "flex" their sizes, either growing to fill unused space or shrinking to avoid overflowing the parent. It is widely used in almost all web applications. & Homepage \\ 
    Grid & CSS & A two-dimensional grid-based layout system. The children of a grid container can be positioned into arbitrary slots in a predefined flexible or fixed-size layout grid. It's more powerful than flexbox and also widely used in web applications. & Homepage \\ 
    Float & CSS & It was originally designed for mixed text and image layout scenarios (text wrapped around images), and was used to implement multi-column layouts in the early days. Its layout function can be replaced by Flexbox or Grid. & Homepage \\ 
    Color & CSS & CSS Color, the basis of page design, involves color definition, functions, etc. & Utility \\ 
    Selector & CSS & Patterns that match against elements in a tree, and as such form one of several technologies that can be used to select nodes in a document. & Utility \\ 
    BOM & Javascript & BOM (Browser Object Model) is the core part of JavaScript on the web. BOM contains many modern browser features, like Window, Location, History, Navigator, Timer, Dialog and more. & Doc \\ 
    DOM & Javascript & DOM (Document Object Model) defines a platform-neutral model for events, aborting activities, and node trees. The main technology to manipulate web page elements. & IDE \\ 
    DOM1 & Javascript & Single-file version of DOM project. The impact of the number of user-compared files and reference relationships on code generation for large models. & IDE \\ 
    Form & Javascript & A form is a component of a web page that has form controls, such as text, buttons, checkboxes, range, or color picker controls. A user can interact with such a form, providing data that can then be sent to the server for further processing (e.g. returning the results of a search or calculation). & Survey \\ 
    ESModule & Javascript & ECMAScript Module System. ECMAScript is Javascript Language Standard, ensures the interoperability of web pages across different web browsers. & Utility \\ 
    Canvas & Javascript & Drawing graphics via JavaScript and the HTML <canvas> element. It can be used for animation, game graphics, data visualization, photo manipulation, and real-time video processing. & Game \\ 
    SVG & Javascript & Scalable Vector Graphics. A XML-based language for describing two-dimensional graphics. SVG drawings can be interactive and dynamic. Animations can be defined and triggered either declaratively. Widely used in data visualization (including chart library) and graphics processing. & Graph \\ 
    SVG-Solar & Javascript & A two-dimensional solar system simulation system created based on SVG technology. & Game \\ 
    SVG-Chart & Javascript & A chart library based on SVG technology, including common charts: LineChart, ScatterChart, StepChart, AreaChart, BarChart, PieChart. & Chart \\ 
    Typescript & Javascript & A strongly typed programming language that builds on JavaScript. TypeScript code converts to JavaScript, which runs anywhere JavaScript runs. Most modern web applications are now built with TypeScript. & Utility \\ 
    Survey & Javascript & A minimal survey system including a survey designer and preview page. including types of questions: Single-Selection Question, Multiple-Selection Question, Open-Ended Question, NPS Question, Likert Question and more. & Survey \\ 
    Draw & Javascript & A minimal draw application with typical basic shapes (Line, Ellipse, Rectangle) and actions (Move, Rotate, Zoom, Delete, Copy, Fill). & Graph \\ 
    ChartBuilder & Javascript & A Chart Builder based on SVG-Chart Project, to generate a chart with user's config. & Chart \\     
  \bottomrule 
  \end{tabular}
  }
  \label{tab:projects-standard}
\end{table*}

\begin{table*}[ht]
  \centering
  \caption{Projects for Web Frameworks Part 1 with GitHub Stars collected in March 2025}
  {\small
  \begin{tabular}{p{1.5cm} l c p{9cm} l}
    \toprule 
    Project & Domain & Stars/K & Description & Scenario \\ 
    \midrule 
    React & UI & 229 & Developers can build user interfaces with components. A component is a piece of the user interface that has its own logic and appearance. A component can be as small as a button, or as large as an entire page. Vue, Angular are the similar frameworks. React has also become one of the preferred UI frameworks in software engineering practice. & Blog \\ 
    Vue & UI & 208 & Developers create reusable components with a template syntax that combines HTML with component logic, and features a reactivity system that automatically keeps the DOM in sync with the application state. & Blog \\ 
    Angular & UI & 96 & Developers create powerful components with dependency injection, and then organize them into modules to build enterprise-level applications.With its change detection mechanism and Ivy rendering engine, it efficiently updates and renders components when data changes, resulting in high-performance applications. & Blog \\ 
    Svelte & UI & 79 & Svelte takes a unique approach by compiling components into efficient vanilla JavaScript at build time, rather than interpreting code at runtime like other frameworks.With its reactive system, it automatically updates the DOM without a virtual DOM, resulting in smaller bundle sizes and better runtime performance. & Blog \\ 
    Threejs & UI & 103 & Developers can harness the powerful capabilities of WebGL without needing to delve into its underlying complexities. It offers a rich set of tools and features, including geometries, materials, lighting, shadows, and animations, making the creation of realistic 3D scenes and interactive experiences more intuitive and convenient. & Game \\ 
    React-no-ts & UI & 229 & A vanilla javascript version of react project based on React project. & Blog \\ 
    Tailwind CSS & CSS & 86 & A utility-first CSS framework packed with classes like flex, pt-4, text-center and rotate-90 that can be composed to build any design, directly in the markup. After the rise of LLM applications in the programming field, Tailwind has continued to attract attention for its simplicity and strong expressiveness, and the community activity has continued to increase. & Homepage \\ 
    Unocss & CSS & 17 & The instant on-demand atomic CSS engine. & Homepage \\ 
    \tiny Styled Components & CSS & 40 & Visual primitives for the component age. Use the best bits of ES6 and CSS to style your apps without stress & Blog \\ 
    Less & CSS & 17 & Less (which stands for Leaner Style Sheets) is a backwards-compatible language extension for CSS. & Survey \\ 
    Sass & CSS & 15 & Provide programming language features to extend CSS, like variables, mixins, nesting, operations, functions and modules. LESS and Stylus are two other similar libraries. SASS (along with LESS and Stylus) is frequently used in large web projects, mainly because it can bring better collaboration efficiency (such as encapsulation into modules and flexible variables). & Survey \\ 
    Stylus & CSS & 11 & An expressive, robust, feature-rich CSS language built for Node.js & Survey \\ 
    Next.js & Fullstack & 127 & A React framework for building fullstack web applications. Use React Components to build user interfaces with a bunch of features out of the box, like Server-side rendering, Static site generation, Routing system and more. Use its API Routes (serverless functions) to handle database queries, authentication, form submissions and any backend logic.  & Shopping \\ 
    Nuxt & Fullstack & 56.7 & A powerful Vue framework that enables features like server-side rendering and static site generation for modern web applications. & Shopping \\ 
    Express.js & Fullstack & 65 & A minimal and flexible Node.js web application framework that provides a robust set of features for web and mobile applications. & Shopping \\ 
    Fastify & Fullstack & 33.4 & A high-performance, low overhead web framework for Node.js. It's designed with performance in mind and is one of the fastest web frameworks available for Node.js. & Shopping \\ 
    Fastify-React & Fullstack & 33 & Fastify implements the backend interface and React implements pure frontend logic. & Shopping \\ 
  \bottomrule 
  \end{tabular}
  }
  \label{tab:projects-framework}
\end{table*}

\begin{table*}[ht]
  \centering
  \caption{Projects for Web Frameworks Part 2 with GitHub Stars collected in March 2025}
  {\small
  \begin{tabular}{llcp{9cm}l}
    \toprule
    Project & Domain & Stars/K & Description & Scenario \\
    \midrule
    Sequelize & Database & 29.8 & A powerful Object-Relational Mapping (ORM) library for Node.js that provides a high-level abstraction for interacting with relational databases like PostgreSQL, MySQL, SQLite, and MSSQL. & Shopping \\ 
    Nosql & Database & 26.8 & A popular NoSQL database that provides high performance, high availability, and easy scalability for modern applications. & Shopping \\ 
    Lowdb & Database & 22 & A minimalist database based on JSON. & Shopping \\ 
    Prisma & Database & 41.8 & Developers can work with databases with intuitive data model, automated migrations, type-safety \& auto-completion. & Shopping \\ 
    Vite & BuildTool & 68 & It provides: A dev server that provides rich feature enhancements to native ES modules, for example Hot Module Replacement (HMR). A build command that bundles your code with Rollup, pre-configured to output optimized static assets for production. An efficient build tool widely used by the community. & Utility \\ 
    Webpack & BuildTool & 64 & A static module bundler for modern JavaScript applications. It internally builds a dependency graph from one or more entry points and then combines every module your project needs into one or more bundles, which are static assets to serve your content from. For a long time, Webpack has been the most widely used build tool in the open source community. In recent years, it has also been challenged by Configless build tools represented by Parcel and Vite, and it is also constantly evolving. & Utility \\ 
    Parcel & BuildTool & 43 & The zero configuration build tool for the web. Parcel combines a great out-of-the-box development experience with a scalable architecture that can take your project from just getting started to massive production application. & Utility \\ 
    Redux & State & 61.1 & Provide a centralized store for state that needs to be used across your entire application, with rules ensuring that state can only be updated in a predictable fashion. With its unidirectional data flow, it makes it easier to understand how data changes in your application, leading to more maintainable code and easier debugging. & Blog \\ 
    Jotai & State & 19.6 & Jotai is a primitive and flexible state management library for React applications. Its significance stems from its atom-based approach, minimal API, and ability to provide granular reactivity without the boilerplate. & Blog \\ 
    Mobx & State & 27.7 & MobX is a battle-tested state management library that makes it simple to connect the reactive data of your application with the UI.Its significance stems from its reactive programming model, observable-based approach, and ability to provide automatic tracking of state dependencies. & Blog \\ 
    Zustand & State & 50.5 & It offers a simple yet powerful store that is unopinionated and can be used with any UI framework, with a focus on simplicity and ease of use. With its subscriber pattern and hooks-first approach, it makes state management straightforward while still offering advanced capabilities like middleware and transient updates. Compared with Redux, Zustand manages the same amount of state with less code and is more concise, and seems to be more suitable for LLM code generation. The test data of this benchmark also supports this view. & Blog \\ 
    Calculator & Other & - & The first project in web-bench. A calculator with basic and science computing features. & Utility \\ 
    \tiny Calculator-files & Other & - & Multiple-files version of Calculator project.  & Utility \\ 
    \tiny Expression Editor & Other & - & A basic expression editor with features like: parenthesis auto closing, syntax highlighting, lint, autocompletion and more. & Sheet \\     
  \bottomrule 
  \end{tabular}
  }
  \label{tab:projects-framework-2}
\end{table*}

\clearpage
\appendix

\section{Appendix: Projects}
\label{appendix:a}


\subsection{Overview}
\label{appendix:a-overview}

See details in Table \ref{tab:projects-standard} \ref{tab:projects-framework} \ref{tab:projects-framework-2}.

\subsubsection{UI Framework}
\label{appendix:a-ui}
UI frameworks offer component-based development, which can improve the efficiency of developing complicated web frontend application.
\paragraph{}
We have four projects: React, Vue, Angular, and Svelte, which represent four popular UI frameworks. These projects share the same tasks, and here is the pass@2 (Best of Five) for these projects:
\noindent
\begin{tabular}{l@{\hskip 6pt}c@{\hskip 6pt}c@{\hskip 6pt}c@{\hskip 6pt}c}
  \toprule
  \textbf{Model} & \textbf{React} & \textbf{Vue} & \textbf{Angular} & \textbf{Svelte} \\
  \midrule
  Claude-3.7-T \footnote{Claude3.7-T is claude-3-7-sonnet-20250219-thinking} & \textbf{60} & 40 & 50 & 55 \\
  Claude-3.7 \footnote{Claude3.7 is claude-3-7-sonnet-20250219} & \textbf{65} & 30 & 40 & 25 \\
  Doubao-1.5-T \footnote{Doubao1.5-T is doubao-pro-1.5-thinking} & \textbf{50} & 40 & 25 & 40 \\
  Doubao-1.5 \footnote{Doubao1.5 is doubao-pro-1.5-32k} & \textbf{35} & \textbf{35} & 5 & 10 \\
  GPT-4o & \textbf{35} & 30 & 5 & 20 \\
  DeepSeek-R1 & \textbf{40} & 30 & 30 & \textbf{40} \\
  \bottomrule
\end{tabular}
\\

\paragraph{Analysis}
The following presents an analysis of these evaluation results: \\
1. When compared to Vue (208k stars), Angular (96k stars), and Svelte (79k stars), React (235k stars) is the most widely-used UI framework in the web domain and provides the largest dataset for LLM training.\\
2. React supports the JSX language feature, which combines JavaScript and HTML, thereby enhancing data density.\\
3. Vue/Svelte exhibits similarities to HTML grammar and demonstrates superior performance in certain models.\\
4. Angular features the most complex grammar, making it difficult for LLMs to learn and utilize.\\

\subsubsection{State Management}
\label{appendix:a-state}
State management framework provides a structured and efficient approach for managing state changes.
\paragraph{}
We have four projects: Redux, Mobx, Zustand, and Jotai, which represent four popular state management frameworks. These projects share the same tasks, and here is the pass@2 (Best of Five) for these projects:

\noindent
\begin{tabular}{l@{\hskip 6pt}c@{\hskip 6pt}c@{\hskip 6pt}c@{\hskip 6pt}c}
  \toprule
  \textbf{Model} & \textbf{Redux} & \textbf{Mobx} & \textbf{Zustand} & \textbf{Jotai} \\
  \midrule
  Claude-3.7-T & 45 & 50 & \textbf{55} & 45 \\
  Claude-3.7 & 45 & \textbf{50} & 45 & 40 \\
  Doubao-1.5-T & \textbf{50} & \textbf{50} & \textbf{50} & 20 \\
  Doubao-1.5 & \textbf{35} & 15 & \textbf{35} & 20 \\
  GPT-4o & 30 & 30 & \textbf{35} & 20 \\
  DeepSeek-R1 & 45 & 40 & \textbf{55} & 40 \\
  \bottomrule 
\end{tabular}

\paragraph{Analysis}
The following presents an analysis of these evaluation results:\\
1. Zustand demonstrates superior performance compared to other state-management frameworks, which can be attributed to its popularity and simple API design.\\
2. Mobx is easy to use, but also easy to make mistakes. For example, Mobx users may forget to add the observer higher-order component (HOC).\\
3. Zustand, Jotai, Mobx features a clean and straightforward API design.\\

\subsubsection{Fullstack Framework}
\label{appendix:a-fullstack}
Fullstack Framework empowers developers to construct both the front-end and back-end components of a web application.
\paragraph{}
There are five projects available to demonstrate the large language model's (LLM) ability to implement fullstack applications. These projects encompass Nextjs, Nuxt, Expressjs, Fastify, and FastifyReact. These projects share the same tasks, and here is the pass@2 (Best of Five) for these projects:

\noindent
\begin{tabular}{l@{\hskip 6pt}c@{\hskip 4pt}c@{\hskip 4pt}c@{\hskip 4pt}c@{\hskip 4pt}c}
  \toprule
  \textbf{Model} & {\small \textbf{Next}} & {\small \textbf{Nuxt}} & {\small\textbf{Express}} & {\small\textbf{Fastify}} & {\small\textbf{F$^*$12}\footnote{F$^*$12 is FastifyReact.}} \\
  \midrule
  Claude-3.7-T & 35 & 25 & \textbf{40} & \textbf{40} & 35 \\
  Claude-3.7 & 40 & 25 & \textbf{45} & 40 & 35 \\
  Doubao-1.5-T & 20 & 20 & 35 & \textbf{45} & 35 \\
  Doubao-1.5 & 15 & 20 & \textbf{50} & 15 & 5 \\
  GPT-4o & 10 & 25 & \textbf{40} & 20 & 5 \\
  DeepSeek-R1 & 20 & \textbf{25} & \textbf{25} & 20 & \textbf{25} \\
  \bottomrule 
\end{tabular}

\paragraph{Analysis}
The following presents an analysis of these evaluation results:\\
1. Expressjs has the longest history (since 2010) in the fullstack domain, and its API is clean and stable across versions.\\
2. Although Nextjs has the highest number of stars, its API is not stable across different versions, and large language models (LLMs) may be confused by the differences between Server Components and Client Components.\\
3. When comparing Next and Nuxt: React-based Next is better than Vue-based Nuxt.\\
4. Fastify-React exhibits poor performance in simple tasks: React Router v7 in Fastify-React is a breaking change version update compared with v6, and in this case, react-router-dom is not required, yet LLMs often overlook this fact.

\subsubsection{Database}
\label{appendix:a-database}
The database is crucial for the backend logic in a fullstack application. Developers can access the database through various Object-Relational Mapping (ORM) tools or Database APIs. 
\paragraph{}
We present four projects (Sequelize, NoSQL, Prisma, and Lowdb) to showcase the Large Language Model's (LLM) capability to utilize ORM/Database APIs. These projects share the same tasks, and here is the pass@2 (Best of Five) for these projects:

\noindent
\begin{tabular}{l@{\hskip 6pt}c@{\hskip 6pt}c@{\hskip 6pt}c@{\hskip 6pt}c}
  \toprule
  \textbf{Model} & {\small\textbf{Sequelize}} & {\small\textbf{NoSQL}} & {\small\textbf{Prisma}} & {\small\textbf{Lowdb}} \\
  \midrule
  Claude-3.7-T & \textbf{40} & 35 & 10 & 35 \\
  Claude-3.7 & 40 & 40 & 45 & \textbf{50} \\
  Doubao-1.5-T & 20 & \textbf{35} & 25 & 30 \\
  Doubao-1.5 & 25 & 20 & 25 & \textbf{30} \\
  GPT-4o & 20 & \textbf{25} & 5 & \textbf{25} \\
  DeepSeek-R1 & \textbf{40} & 35 & 20 & 30 \\
  \bottomrule 
\end{tabular}

\paragraph{Analysis}
The following presents an analysis of these evaluation results:\\
1. Nosql exhibits good performance. The Document Model of MongoDB is significantly easier to use than that of relation-based databases, and it is also suitable for Node.js.\\
2. Lowdb is an extremely simple JSON database. With lowdb, database APIs are the same as JavaScript Array APIs, which is simple enough for LLM to understand.\\

\subsection{Flex}
\paragraph{Definition}
from \href{https://www.w3.org/TR/css-flexbox-1/#abstract}{W3C}:
\begin{quote}
The specification describes a CSS box model optimized for user interface design. In the flex layout model, the children of a flex container can be laid out in any direction, and can "flex" their sizes, either growing to fill unused space or shrinking to avoid overflowing the parent. Both horizontal and vertical alignment of the children can be easily manipulated. Nesting of these boxes (horizontal inside vertical, or vertical inside horizontal) can be used to build layouts in two dimensions.
\end{quote}

from \href{https://developer.mozilla.org/en-US/docs/Web/CSS/CSS_flexible_box_layout}{MDN}:
\begin{quote}
The CSS flexible box layout module defines a CSS box model optimized for user interface design, and the layout of items in one dimension. In the flex layout model, the children of a flex container can be laid out in any direction, and can "flex" their sizes, either growing to fill unused space or shrinking to avoid overflowing the parent. Both horizontal and vertical alignment of the children can be easily manipulated.
\end{quote}

\paragraph{Feature Coverage} 
This table shows the features used in solution codes ('src/') of the project. This is roughly the feature set that LLM will cover. Appendix \ref{appendix:a} unless otherwise specified, the subsequent chapters have the same meaning.

\begin{tabular}{lclc}
\toprule
Feature &  & Feature &  \\ 
\midrule
display:flex & \checkmark & column-gap & \checkmark \\ 
flex-direction & \checkmark & gap & \checkmark \\ 
flex-wrap & \checkmark & order & \checkmark \\ 
flex-flow & \checkmark & flex-grow & \checkmark \\ 
justify-content & \checkmark & flex-shrink & \checkmark \\ 
align-items & \checkmark & flex-basis & \checkmark \\ 
align-content & \checkmark & flex & \checkmark \\ 
row-gap & \checkmark & align-self & \checkmark \\ 
\bottomrule 
\end{tabular}

\subsection{Grid}
\paragraph{Definition}
from \href{https://www.w3.org/TR/css-grid-1/#abstract}{W3C}:
\begin{quote}
  This CSS module defines a two-dimensional grid-based layout system, optimized for user interface design. In the grid layout model, the children of a grid container can be positioned into arbitrary slots in a predefined flexible or fixed-size layout grid.
\end{quote}

from \href{https://developer.mozilla.org/en-US/docs/Learn/CSS/CSS_layout/Grids}{MDN}:
\begin{quote}
  CSS grid layout is a two-dimensional layout system for the web. It lets you organize content into rows and columns and offers many features to simplify the creation of complex layouts.
\end{quote}

\paragraph{Feature Coverage}
This table shows the features used in solution codes ('src/') of the project.

\noindent
\begin{tabular}{lclc}
  \toprule
  Feature &  & Feature &  \\
  \midrule
  display:grid & \checkmark & grid-auto-rows & \checkmark \\
  \scriptsize grid-template-columns & \checkmark & \scriptsize grid-column-end & \checkmark \\
  grid-template-rows & \checkmark & grid-row-start & \checkmark \\
  grid-template-areas & \checkmark & grid-row-end & \checkmark \\
  grid-template & \checkmark & grid-column & \checkmark \\
  column-gap & \checkmark & grid-row & \checkmark \\
  row-gap & \checkmark & grid-area & \checkmark \\
  gap & \checkmark & justify-self & \texttimes \\
  justify-items & \checkmark & align-self & \texttimes \\
  align-items & \checkmark & place-self & \texttimes \\
  place-items & \checkmark & order & \checkmark \\
  justify-content & \checkmark & min-content & \checkmark \\
  align-content & \texttimes & max-content & \checkmark \\
  place-content & \checkmark & fit-content() & \texttimes \\
  grid-auto-columns & \texttimes & minmax() & \texttimes \\
  grid-auto-flow & \checkmark & min() & \checkmark \\
  grid & \checkmark & max() & \checkmark \\
  grid-column-start & \checkmark & subgrid & \texttimes \\
  \bottomrule 
\end{tabular}

\subsection{CSS Selector}
\paragraph{Definition}
from \href{https://drafts.csswg.org/selectors/}{W3C}:
\begin{quote}
Selectors are patterns that match against elements in a tree, and as such form one of several technologies that can be used to select nodes in a document.
They are a core component of CSS (Cascading Style Sheets), which uses Selectors to bind style properties to elements in the document.
\end{quote}

\paragraph{Project Design}
We a page with complex DOM structure to test different types of selectors and specificities. Tasks keywords:

\noindent
\begin{tabular}{ll}
  \toprule
  Task & Task \\
  \midrule
  1. \small Type/Class/ID Selector & 11. Specificity \\
  2. Compound Selector & 12. Specificity \\
  3. Selector list & 13. Specificity \\
  4. Combinator & 14. \scriptsize Pseudo Class, Form \\
  5. Combinator & 15. \scriptsize Pseudo Class, Link \\
  6. Pseudo Class & 16. Nested Structure \\
  7. Nesting Selector & 17. :is() \\
  8. Pseudo Class & 18. :is() Deep Levels \\
  9. Pseudo Class & 19. :is() Deep Levels \\
  10. Pseudo Element & 20. :is() Deep Levels \\
  \bottomrule 
\end{tabular}

\paragraph{Feature Coverage}
This table shows the features used in solution codes ('src/') of the project.

\noindent
\begin{tabular}{lclc}
  \toprule
  Feature &  & Feature &  \\
  \midrule
  * & \checkmark & :focus & \checkmark \\
  E & \checkmark & :focus-within & \checkmark \\
  \#id & \checkmark & :enabled & \checkmark \\
  .class & \checkmark & :disabled & \checkmark \\
  E.class1.class2 & \checkmark & :default & \texttimes \\
  sel1, sel2 & \checkmark & :checked & \texttimes \\
  E F & \checkmark & :valid & \checkmark \\
  E > F & \checkmark & :invalid & \checkmark \\
  E + F & \checkmark & :required & \checkmark \\
  E ~ F & \checkmark & :optional & \checkmark \\
  E[foo] & \checkmark & :root & \texttimes \\
  E[foo="bar"] & \checkmark & :empty & \checkmark \\
  E[foo\texttt{\~=}"bar"] & \checkmark & :nth-child(n) & \checkmark \\
  E[foo\texttt{\^=}"bar"] & \checkmark & \small :nth-last-child(n) & \checkmark \\
  E[foo\$="bar"] & \checkmark & :only-child & \texttimes \\
  E[foo*="bar"] & \checkmark & ::before & \checkmark \\
  E[foo\texttt{|=}"bar"] & \checkmark & ::after & \checkmark \\
  :link & \checkmark & :is() & \checkmark \\
  :visited & \checkmark & :where() & \texttimes \\
  :target & \checkmark & :has() & \checkmark \\
  :active & \checkmark & :not() & \checkmark \\
  :hover & \checkmark & &  \\
  \bottomrule 
\end{tabular}

\subsection{DOM}
\paragraph{Definition}
from \href{https://dom.spec.whatwg.org/#introduction-to-the-dom}{W3C}:
\begin{quote}
  In its original sense, "The DOM" is an API for accessing and manipulating documents (in particular, HTML and XML documents). Each such document is represented as a node tree. Some of the nodes in a tree can have children, while others are always leaves.
\end{quote}

from \href{https://developer.mozilla.org/en-US/docs/Web/API/Document_Object_Model}{MDN}:
\begin{quote}
  The Document Object Model ( DOM ) connects web pages to scripts or programming languages by representing the structure of a document—such as the HTML representing a web page—in memory.
\end{quote}

\begin{figure}[htbp!]
  \centering
  \includegraphics[width=1\linewidth]{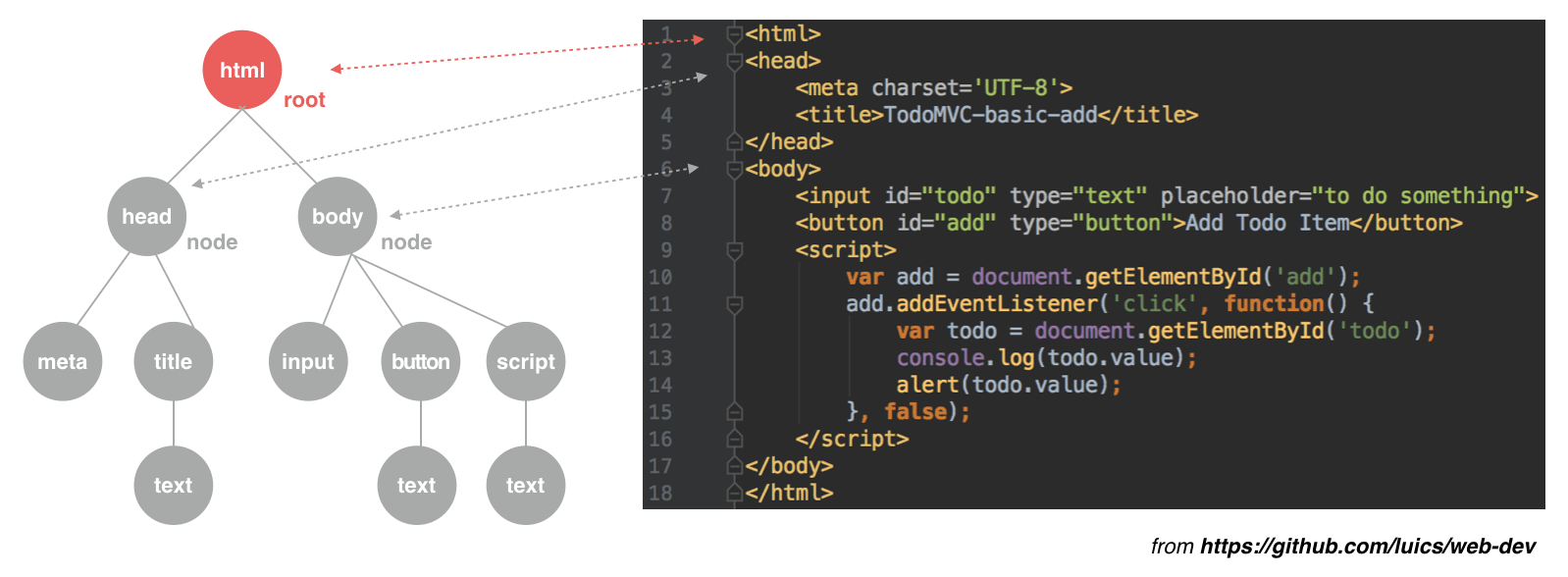}
  \caption{DOM Tree \& HTML}
  \label{fig:dom}
\end{figure}

\paragraph{Project Design}
We have developed a minimal File Explorer. In the project, there are 2 entry types including text file and directory. Typical File Explorer features are: Display Hierarchy Tree, Create, Delete, Rename, Modify (file only), Move, Copy and more. Tasks keywords:
\begin{lstlisting}[style=markdown]
1. Tools
2. Add File
3. Add Dir
4. SelectedEntry
5. Delete Entry
6. Drag and move
7. Context Menu, add/delete Entry, for .entry
8. Context Menu, add/delete Entry, for .entries
9. Context Menu, copy and paste
10. Context Menu, rename
11. Import
12. Export
13. Modify file content
14. Shortcuts, copy/paste/add
15. Shortcuts, delete
16. Generate large entry list
17. Resizer
18. Drag resizer
19. Filter
20. Context Menu, cut
\end{lstlisting}

\paragraph{Feature Coverage}
This table shows the features used in solution codes ('src/') of the project.

\begin{figure}[htbp!]
  \centering
  \includegraphics[width=1\linewidth]{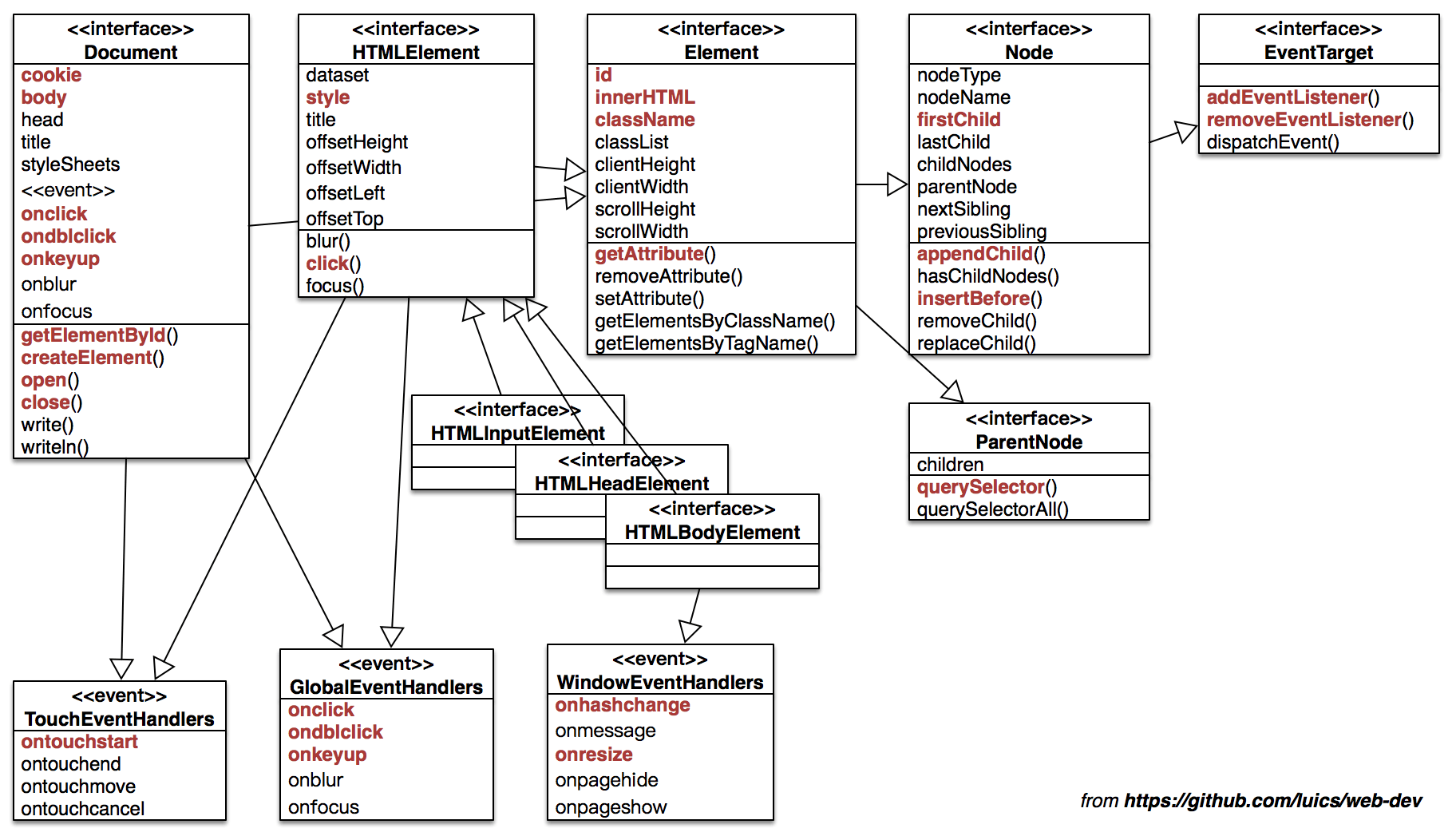}
  \caption{DOM Class Hierarchy (Partially)}
  \label{fig:dom-hierarchy}
\end{figure}

\begin{tabular}{lclc}
  \toprule
  Feature &  \\
  \midrule
  Document.getElementById() & \checkmark \\
  Document.createElement() & \checkmark \\
  Document.createEvent() & \texttimes \\
  ParentNode.querySelector() & \checkmark \\
  ParentNode.querySelectorAll() & \checkmark \\
  ParentNode.children & \checkmark \\
  EventTarget.addEventListener() & \checkmark \\
  EventTarget.removeEventListener() & \texttimes \\
  Node.appendChild() & \checkmark \\
  Node.insertBefore() & \checkmark \\
  Node.removeChild() & \checkmark \\
  Node.replaceChild() & \checkmark \\
  Node.hasChildNodes() & \texttimes \\
  Node.textContent & \checkmark \\
  Node.firstChild & \checkmark \\
  Node.lastChild & \checkmark \\
  Node.childNodes & \checkmark \\
  Node.parentNode & \checkmark \\
  Node.nextSibling & \checkmark \\
  Node.previousSibling & \checkmark \\
  Element.id & \checkmark \\
  Element.innerHTML & \checkmark \\
  Element.className & \checkmark \\
  Element.classList & \checkmark \\
  Element.clientHeight & \texttimes \\
  Element.clientWidth & \texttimes \\
  Element.scrollHeight & \texttimes \\
  Element.scrollWidth & \texttimes \\
  Element.getAttribute() & \checkmark \\
  Element.setAttribute() & \checkmark \\
  Element.hasAttribute() & \checkmark \\
  Element.removeAttribute() & \texttimes \\
  Element.getElementsByClassName() & \texttimes \\
  Element.getElementsByTagName() & \texttimes \\
  Element.after() & \checkmark \\
  Element.before() & \texttimes \\
  Element.remove() & \checkmark \\
  Element.closest() & \checkmark \\
  \bottomrule
\end{tabular}

\subsection{SVG}
\paragraph{Definition}
from \href{https://svgwg.org/svg2-draft/Overview.html}{W3C}:
\begin{quote}
  SVG is a language for describing two-dimensional graphics.
SVG drawings can be interactive and dynamic. Animations can be defined and triggered either declaratively (i.e., by embedding SVG animation elements in SVG content) or via scripting.
\end{quote}

from \href{https://developer.mozilla.org/en-US/docs/Web/SVG}{MDN}:
\begin{quote}
  Scalable Vector Graphics (SVG) is an XML-based markup language for describing two-dimensional based vector graphics.
  As such, it's a text-based, open Web standard for describing images that can be rendered cleanly at any size and are designed specifically to work well with other web standards including CSS, DOM, JavaScript, and SMIL. SVG is, essentially, to graphics what HTML is to text.
\end{quote}

\paragraph{Project Design}
We have developed a minimal draw application. Typical draw features are: \\
1. Shapes: Line, Rectangle, Ellipse, Circle, Polyline, Polygon, Path, Text \\
2. Common Property: LineWidth, Color \\
3. Operations: Move, Rotate, Zoom, Delete, Copy, Fill \\

\noindent
Task keywords:
\begin{lstlisting}[style=markdown]
1. Draw Line
2. Draw Rect
3. Draw Circle
4. Draw Ellipse
5. Delete Shape
6. Fill Shape
7. Copy Shape
8. Minimal Line and Rect
9. Minimal Circle and Ellipse
10. Draw Triangle
11. Draw Trapezoid
12. Draw Hexagon
13. Draw Curve (Path)
14. Draw Polyline
15. Draw Text
16. Move Shape
17. Shortcuts
18. Rotate Shape
19. Zoom Shape
20. Combined Action
\end{lstlisting}

\paragraph{Feature Coverage}
This table shows the features used in solution codes ('src/') of the project.

{\small

\noindent
\begin{tabular}{lclc}
  \toprule
  Feature &  & Feature &  \\
  \midrule
  Shapes \& Path & - & $<$animate$>$ & \texttimes \\
  $<$rect$>$ & \checkmark & $<$animateMotion$>$ & \checkmark \\
  $<$circle$>$ & \checkmark &   $<$mpath$>$ & \checkmark \\
  $<$ellipse$>$ & \checkmark & \tiny $<$animateTransform$>$ & \checkmark \\
  $<$line$>$ & \checkmark & $<$set$>$ & \texttimes \\
  $<$polyline$>$ & \checkmark & Styles & - \\
  $<$polygon$>$ & \checkmark & $<$linearGradient$>$ & \checkmark \\
  $<$path$>$ & \checkmark & $<$radialGradient$>$ & \checkmark \\
  Text & - & $<$pattern$>$ & \texttimes \\
  $<$text$>$ & \checkmark & $<$marker$>$ & \checkmark \\
  $<$tspan$>$ & \texttimes & $<$filter$>$ & \texttimes \\
  $<$textPath$>$ & \texttimes & $<$clipPath$>$ & \checkmark \\
  \tiny Document Structure & - & $<$mask$>$ & \checkmark \\
  $<$svg$>$ & \checkmark & color & \checkmark \\
  $<$g$>$ & \checkmark & transform & \checkmark \\
  $<$defs$>$ & \checkmark & transform-origin & \checkmark \\
  $<$symbol$>$ & \checkmark & fill & \checkmark \\
  $<$use$>$ & \checkmark & fill-rule & \texttimes \\
  $<$title$>$ & \checkmark & fill-opacity & \checkmark \\
  $<$desc$>$ & \texttimes & stroke & \checkmark \\
  $<$switch$>$ & \texttimes & stroke-width & \checkmark \\
  $<$a$>$ & \texttimes & stroke-linecap & \checkmark \\
  $<$view$>$ & \texttimes & stroke-linejoin & \checkmark \\
  \tiny Embedded Content & - & stroke-miterlimit & \texttimes \\
  $<$image$>$ & \checkmark & stroke-dasharray & \checkmark \\
  $<$foreignObject$>$ & \checkmark & stroke-dashoffset & \checkmark \\
  Animation (SMIL) & - & stroke-opacity & \checkmark \\
  \bottomrule 
\end{tabular}
}

\subsection{React}
\paragraph{Definition}
from \href{https://github.com/facebook/react/}{React}:
\begin{quote}
  React is a JavaScript library for building user interfaces. \\
- Declarative: React makes it painless to create interactive UIs. Design simple views for each state in your application, and React will efficiently update and render just the right components when your data changes. Declarative views make your code more predictable, simpler to understand, and easier to debug. \\
- Component-Based: Build encapsulated components that manage their own state, then compose them to make complex UIs. Since component logic is written in JavaScript instead of templates, you can easily pass rich data through your app and keep the state out of the DOM.\\
- Learn Once, Write Anywhere: We don't make assumptions about the rest of your technology stack, so you can develop new features in React without rewriting existing code. React can also render on the server using Node and power mobile apps using React Native.
\end{quote}

\paragraph{Project Design}
This Project create a blog system to test the ability of LLM to use React.js. Tasks is designed to verify Agent's knowledge of React\.js:
\begin{lstlisting}[style=markdown]
1. Create components and use props
2. List Rendering
3. State & Event
4. Visible-Control (Modal)
5. useEffect
6. Form Handling
7. React Context API
8. Custom Hook & Delete in List State
9. Edit in List State & Reuse of CreateForm
10. Filter List & useMemo
11. Virtual List
12. useSyncExternalStore
13. ReactPortal & useRef (Tooltip)
14. XSS prevention
15. ReactDOM API (showToast)
16. TextOverflow
17. use forwardRef and useImperativeHandle
18. write a simple React Router
19. write a complete game
20. Communication between pages.
\end{lstlisting}
\paragraph{Feature Coverage}
This table shows the features used in solution codes ('src/') of the project.
\noindent
\begin{tabular}{lclc}
  \toprule
  Feature &  & Feature &  \\
  \midrule
  Component & \checkmark & useSyncExternalStore & \checkmark \\
  Props & \checkmark & createPortal & \checkmark \\
  State & \checkmark & \scriptsize dangerouslySetInnerHTML & \checkmark \\
  Event & \checkmark & XSS Prevention & \checkmark \\
  List & \checkmark & React DOM API & \checkmark \\
  Form & \checkmark & startTransition & \texttimes \\
  Custom Hook & \checkmark & useDeferredValue & \texttimes \\
  Context API & \checkmark & forwardRef & \checkmark \\
  useEffect & \checkmark & useImperativeHandle & \checkmark \\
  useRef & \checkmark & lazy & \texttimes \\
  useCallback & \checkmark & memo & \texttimes \\
  useMemo & \checkmark &  & \\
  \bottomrule 
\end{tabular}

\subsection{Threejs}
\paragraph{Definition}
from \href{https://github.com/mrdoob/three.js}{Threejs GitHub}:
\begin{quote}
  The aim of the project is to create an easy-to-use, lightweight, cross-browser, general-purpose 3D library. The current builds only include WebGL and WebGPU renderers but SVG and CSS3D renderers are also available as addons.
\end{quote}

from \href{https://threejs.org/manual/#en/fundamentals}{Threejs Manual}:
\begin{quote}
  Three.js is a 3D library that tries to make it as easy as possible to get 3D content on a webpage.
\end{quote}

\paragraph{Project Design}
We have developed a 3D version of the Snake game by threejs. We have added a portal mechanism to this game. Typical Snake game tasks are:
\begin{lstlisting}[style=markdown]
1. Generate scene, Create a renderer and enable auto-refresh
2. Generate an 8 * 8 floor
3. Create a camera, Place a point light source
4. Generate a snake group and snakeHead
5. Generate snakeBody
6. Support movement control in forward, backward, left, and right directions
7. Generate a surrounding fence and add collision detection
8. Generate candy
9. Grow by one unit when the snake eats candy
10. Candy Animation
11. Generate portal
12. Portal animation
13. Candy cannot be generated in the portal
14. Mouse view rotation
15. Click 'h' to center the view
16. View automatically follows the snake's head movement
17. Snake resets to initial state when entering a dead-end
18. Snake auto movement
19. The snake changes color after passing through the portal
20. Game success when the snake's length reaches 20
\end{lstlisting}

\paragraph{Feature Coverage}
This table shows the features used in solution codes ('src/') of the project.

\noindent
\begin{tabular}{lclc}
  \toprule
  Feature &  & Feature &  \\
  \midrule
  Scenes & \checkmark & Objects & \checkmark \\
Cameras & \checkmark & - BatchedMesh & \texttimes \\
- Camera & \checkmark & - Bone & \texttimes \\
- CubeCamera & \texttimes & - Group & \checkmark \\
\scriptsize - OrthographicCamera & \texttimes & - Line & \texttimes \\
\scriptsize - PerspectiveCamera & \checkmark & - Mesh & \checkmark \\
- StereoCamera & \texttimes & - Skeleton & \texttimes \\
Materials & \checkmark & - Sprite & \texttimes \\
Geometry & \checkmark & Light \& Shadow & \checkmark \\
- BoxGeometry & \checkmark & Loader & \texttimes \\
- CircleGeometry & \texttimes & Math & \checkmark \\
- ConeGeometry & \checkmark & Audio & \texttimes \\
- CylinderGeometry & \texttimes & Helper & \texttimes \\
- PlaneGeometry & \checkmark & Animation & \checkmark \\
\scriptsize - PolyhedronGeometry & \texttimes & - AnimationAction & \checkmark \\
- RingGeometry & \checkmark & - AnimationClip & \checkmark \\
- ShapeGeometry & \texttimes & - AnimationMixer & \checkmark \\
- SphereGeometry & \checkmark & - \tiny AnimationObjectGroup & \texttimes \\
Textures & \texttimes & - AnimationUtils & \texttimes \\
Renderers & \checkmark & &  \\
  \bottomrule 
\end{tabular}

\subsection{Redux}
\paragraph{Definition}
from \href{https://redux.js.org/introduction/getting-started}{Redux}:
\begin{quote}
  Redux is a JS library for predictable and maintainable global state management.
It helps you write applications that behave consistently, run in different environments (client, server, and native), and are easy to test. On top of that, it provides a great developer experience, such as live code editing combined with a time traveling debugger.
\end{quote}

\paragraph{Feature Coverage}
This table shows the features used in solution codes ('src/') of the project.

\noindent
\begin{tabular}{lclc}
  \toprule
  Feature &  & Feature &  \\
  \midrule
  Store & \checkmark & \scriptsize createEntityAdapter & \texttimes \\
Actions & \checkmark & Redux DevTools & \texttimes \\
Reducers & \checkmark & Immutability & \checkmark \\
Middleware & \checkmark & React-Redux & \checkmark \\
Combine Reducers & \checkmark & connect & \texttimes \\
Async Actions & \checkmark & useSelector & \checkmark \\
Redux Toolkit & \checkmark & useDispatch & \checkmark \\
createSlice & \checkmark & Provider & \checkmark \\
configureStore & \checkmark & Redux Thunk & \checkmark \\
createAsyncThunk & \checkmark & Redux Saga & \texttimes \\
createSelector & \texttimes & Redux Observable & \texttimes \\
  \bottomrule 
\end{tabular}

\subsection{Zustand}
\paragraph{Definition}
from \href{https://zustand.docs.pmnd.rs/getting-started/introduction}{Zustand}:
\begin{quote}
  A small, fast, and scalable bearbones state management solution. Zustand has a comfy API based on hooks. It isn't boilerplated or opinionated, but has enough convention to be explicit and flux-like.
\end{quote}

\paragraph{Feature Coverage}
This table shows the features used in solution codes ('src/') of the project.

\noindent
\begin{tabular}{lclc}
  \toprule
  Feature &  & Feature &  \\
  \midrule
  Store & \checkmark & Immer Integration & \checkmark \\
  State Getters & \checkmark & Immutability & \checkmark \\
  State Setters & \checkmark & React Hooks & \checkmark \\
  Combine Stores & \texttimes & useStore & \checkmark \\
  Async Actions & \checkmark & useShallow & \checkmark \\
  Selectors & \texttimes & DevTools & \texttimes \\
  Computed Values & \checkmark & Transient Updates & \texttimes \\
  Persistence & \texttimes & Optics & \texttimes \\
  \bottomrule 
\end{tabular}

\subsection{Survey}
\label{appendix:a-survey}
\paragraph{Project Design}
We have developed a minimal survey system including a survey designer and preview page. Task keywords are:
\begin{lstlisting}[style=markdown]
1. Question
2. Design Page, CRUD, Save Data
3. Preview Page, Render Form
4. Single-Selection Question in Design and Preview page
5. Multiple-Selection Question
6. Open-Ended Question (Text)
7. Rating Question (Stars)
8. Ranking Question
9. NPS Question (0-10)
10. Likert Question (Matrix)
11. Question Contents and DnD
12. Validation
13. Validation in Design and Preview
14. OpenQuestion Validation
15. MultiSelectionQuestion Validation
16. DataQuestion and Validation
17. Preview ProgressBar
18. Shuffle Options
19. Branch
20. Branch
\end{lstlisting}

\subsection{SASS}
\paragraph{Definition}
from \href{https://sass-lang.com}{SASS}:
\begin{quote}
  Sass is the most mature, stable, and powerful professional grade CSS extension language in the world.
\end{quote}

\paragraph{Project Design}
Based on project \ref{appendix:a-survey}. This project use SASS syntax to rewrite all the styles.

\paragraph{Feature Coverage}
This table shows the features used in solution codes ('src/') of the project.

\begin{tabular}{lclc}
  \toprule
  Feature &  & Feature &  \\
  \midrule
  Variables & \checkmark & Operations & \checkmark \\
  --Map & \checkmark & Functions & \checkmark \\
  --List & \texttimes & Inheritance & \texttimes \\
  Mixins & \checkmark & Comments & \checkmark \\
  Nesting & \checkmark & Modules & \checkmark \\
  \bottomrule 
\end{tabular}

\subsection{Vite}
\paragraph{Definition}
from \href{https://vite.dev/guide/}{Vite}:
\begin{quote}
  Vite (French word for "quick", pronounced /vit/, like "veet") is a build tool that aims to provide a faster and leaner development experience for modern web projects. It consists of two major parts: \\
  - A dev server that provides rich feature enhancements over native ES modules, for example extremely fast Hot Module Replacement (HMR).\\
  - A build command that bundles your code with Rollup, pre-configured to output highly optimized static assets for production.\\
\end{quote}

\paragraph{Project Design} 
Task Keywords:
\begin{lstlisting}[style=markdown]
1. Initialize project
2. Add path alias
3. Global variable replacement
4. SPA history fallback
5. Browser compatibility
6. Proxy
7. Using images
8. Using CSS/Less/CSS Modules
9. Using TypeScript
10. Using React/Vue
11. Hide sourcemap files
12. Remove all console.log call expressions
13. Extract licenses into standalone file
14. In-memory virtual module
15. Support mock in dev server
16. Compress images
17. Support importing compiled HTML from markdown files
18. Support importing frontmatter from markdown files
19. Handle assets resolving inside markdown files
20. Add i18n feature for markdown file
\end{lstlisting}

\paragraph{Feature Coverage}
This table shows the features used in solution codes ('src/') of the project.

\begin{tabular}{lc}
  \toprule
  Feature  &  \\
  \midrule
  root & \texttimes \\
  base & \checkmark \\
  mode & \texttimes \\
  define & \checkmark \\
  plugins & \checkmark \\
  publicDir & \texttimes \\
  cacheDir & \texttimes \\
  resolve.alias & \checkmark \\
  resolve.dedupe & \texttimes \\
  resolve.conditions & \texttimes \\
  resolve.mainFields & \texttimes \\
  resolve.extensions & \texttimes \\
  resolve.preserveSymlinks & \texttimes \\
  html.cspNonce & \texttimes \\
  css.modules & \texttimes \\
  css.postcss & \texttimes \\
  css.preprocessorOptions & \texttimes \\
  css.pre....additionalData & \texttimes \\
  css.preprocessorMaxWorkers & \texttimes \\
  css.devSourcemap & \texttimes \\
  css.transformer & \texttimes \\
  css.lightningcss & \texttimes \\
  json.namedExports & \texttimes \\
  json.stringify & \texttimes \\
  esbuild & \texttimes \\
  assetsInclude & \texttimes \\
  logLevel & \texttimes \\
  customLogger & \texttimes \\
  clearScreen & \texttimes \\
  envDir & \texttimes \\
  envPrefix & \texttimes \\
  appType & \texttimes \\
  future & \texttimes \\
  \bottomrule 
\end{tabular}
\noindent
\begin{tabular}{lclc}
  \toprule
  Feature &  & Feature &  \\
  \midrule
  server.host & \checkmark & server.hmr & \texttimes \\
  \scriptsize server.allowedHosts & \texttimes & server.warmup & \texttimes \\
  server.port & \checkmark & server.watch & \texttimes \\
  server.strictPort & \texttimes & \scriptsize server.middlewareMode & \texttimes \\
  server.https & \texttimes & server.fs.strict & \texttimes \\
  server.open & \texttimes & server.fs.allow & \texttimes \\
  server.proxy & \checkmark & server.fs.deny & \texttimes \\
  server.cors & \texttimes & server.origin & \texttimes \\
  server.headers & \texttimes & \scriptsize server.sourcemapIgnoreList & \texttimes \\
  \bottomrule 
\end{tabular}
\noindent
\begin{tabular}{lclc}
  \toprule
  Feature &   \\
  \midrule
  build.target & \checkmark \\
  build.modulePreload & \texttimes \\
  build.polyfillModulePreload & \texttimes \\
  build.outDir & \checkmark \\
  build.assetsDir & \texttimes \\
  build.assetsInlineLimit & \texttimes \\
  build.cssCodeSplit & \texttimes \\
  build.cssTarget & \texttimes \\
  build.cssMinify & \texttimes \\
  build.sourcemap & \checkmark \\
  build.rollupOptions & \checkmark \\
  build.commonjsOptions & \texttimes \\
  build.dynamicImportVarsOptions & \texttimes \\
  build.lib & \texttimes \\
  build.manifest & \texttimes \\
  build.ssrManifest & \texttimes \\
  build.ssr & \texttimes \\
  build.emitAssets & \texttimes \\
  build.ssrEmitAssets & \texttimes \\
  build.minify & \texttimes \\
  build.terserOptions & \texttimes \\
  build.write & \texttimes \\
  build.emptyOutDir & \texttimes \\
  build.copyPublicDir & \texttimes \\
  build.reportCompressedSize & \texttimes \\
  build.chunkSizeWarningLimit & \texttimes \\
  build.watch & \texttimes \\
  \bottomrule 
\end{tabular}

\subsection{Nextjs}
\paragraph{Definition}
from \href{https://nextjs.org/docs}{Nextjs}:
\begin{quote}
Next.js is a React framework for building full-stack web applications. You use React Components to build user interfaces, and Next.js for additional features and optimizations.\\
Under the hood, Next.js also abstracts and automatically configures tooling needed for React, like bundling, compiling, and more. This allows you to focus on building your application instead of spending time with configuration.\\
Whether you're an individual developer or part of a larger team, Next.js can help you build interactive, dynamic, and fast React applications.\\
\end{quote}

\paragraph{Project Design}
Tasks are designed to verify Agent's knowledge of Next.js:
\begin{lstlisting}[style=markdown]
1. Basic Router (App Router or Page Router)
2. Layout
3. NotFound + Link
4. Route Handler
5. List Page with Data Fetching from Sqlite
6. Add Dynamic Routes For Detail Page
7. Simple JWT Authentication & Cookie
8. Login Form with Client Component
9. Register Form with Server Actions
10. Static User Profile with Redirect
11. User Widget with REALTIME STATUS (Agent might forget it). Combine React Context Provider & Server Actions.
12. ACCURATE Number Increment with Client Component & Server Actions, update coin displayed STATIC User Profile
13. Add Admin Portals, add middleware to check privileges
14. Wish List
15. Cart Functionality
16. Place Order
17. Pay Order
18. Refunds Order
19. Comment System
20. Invitation System (Bonus when new User Registered or Pay a New Order)
\end{lstlisting}

\paragraph{Feature Coverage}
This table shows the features used in solution codes ('src/') of the project.\\
\noindent
\begin{tabular}{lclc}
  \toprule
  Feature &  & Feature &  \\
  \midrule
  App Router & \checkmark & ISR & \texttimes \\
  Page Router & \checkmark & Server Components & \checkmark \\
  Pages Router & \checkmark & Client Components & \checkmark \\
  Layout & \checkmark & Server Actions & \checkmark \\
  Route Handler & \checkmark & getStaticProps & \texttimes \\
  Redirect & \checkmark & Cookie & \checkmark \\
  NotFoundPage & \checkmark & getServerSideProps & \texttimes \\
  API Routes & \checkmark & Image Optimization & \texttimes \\
  Dynamic Routes & \checkmark & Dynamic Imports & \checkmark \\
  Route Groups & \texttimes & Middleware & \checkmark \\
  Parallel Routes & \texttimes & Loading UI & \texttimes \\
  \scriptsize Intercepting Routes & \texttimes & Error Boundaries & \texttimes \\
Static Generation & \texttimes & generateMetadata & \texttimes \\
SSR & \checkmark & Edge Runtime & \texttimes \\
  \bottomrule 
\end{tabular}

\subsection{Prisma}
\paragraph{Definition}
from \href{https://www.prisma.io/docs/orm/overview/introduction/why-prisma}{Prisma}:
\begin{quote}
  Prisma ORM's main goal is to make application developers more productive when working with databases. Here are a few examples of how Prisma ORM achieves this:\\
  - Thinking in objects instead of mapping relational data\\
  - Queries not classes to avoid complex model objects\\
  - Single source of truth for database and application models\\
  - Healthy constraints that prevent common pitfalls and anti-patterns\\
  - An abstraction that makes the right thing easy ("pit of success")\\
  - Type-safe database queries that can be validated at compile time\\
  - Less boilerplate so developers can focus on the important parts of their app\\
  - Auto-completion in code editors instead of needing to look up documentation\\
\end{quote}

\paragraph{Feature Coverage}
This table shows the features used in solution codes ('src/') of the project.

\begin{tabular}{lclc}
  \toprule
  Feature &   \\
  \midrule
  Schema Definition & \checkmark \\
  Model Creation (Prisma Client) & \checkmark \\
  Document Creation (create) & \checkmark \\
  Document Creation (createMany) & \checkmark \\
  Document Query (findMany) & \checkmark \\
  Document Query (findUnique) & \checkmark \\
  Document Update (update) & \checkmark \\
  Document Update (updateMany) & \checkmark \\
  Document Deletion (delete) & \checkmark \\
  Document Deletion (deleteMany) & \checkmark \\
  Relations (include) & \checkmark \\
  Middlewares & \texttimes \\
  Computed Fields & \texttimes \\
  Indexes & \texttimes \\
  Validation & \texttimes \\
  Aggregation & \texttimes \\
  Transactions (\$transaction) & \checkmark \\
  Extensions & \texttimes \\
  Raw Database Access & \texttimes \\
  Connection Management & \texttimes \\
  Fluent API & \texttimes \\
  Nested Writes & \texttimes \\
  Query Filters & \texttimes \\
  Instance Methods & \texttimes \\
  \bottomrule 
\end{tabular}

\section{Appendix: Web-Agent}
\label{appendix:b}

Web-Agent Workflow is shown in Figure \ref{fig:web-agent-workflow}.

\paragraph{1. Build Prompts}
Using system prompt (SP), task description, files, and error messages. If the combined input exceeds the context length, it will be truncated.\\

\paragraph{System Prompt} \ \\
\begin{lstlisting}[style=markdown]
When generating new code:
1. Always produce a single code block.
2. Never separate the code into multiple code blocks.
3. Only include the code that is being added.
4. No explanation, no issue, only code.
5. Never omit any code.,
6. If the user submits a code block that contains a filename in the language specifier, always include the filename in any code block you generate based on that file. The filename should be on the next line as the language specifier in your code block.
7. Don't repeat filename in code block.

Always follow these guidelines when generating code responses.
\end{lstlisting}

\paragraph{Task Description} \ \\
\begin{lstlisting}[style=markdown]
add divs(class 'header', 'footer', 'content') with arbitrary text in '.root' element. '.root' occupies total viewport and children elements together occupy total '.root' space. header (border-box) is always fixed at the top of '.root'; footer (border-box) is always fixed at the bottom of '.root'; content (border-box) occupies the remaining '.root' space. USE tailwind grid only, NO flex, float and position, NO js.
\end{lstlisting}

\paragraph{Files}  \ \\
\begin{lstlisting}[style=html]
```
index.html
<!DOCTYPE html>
<html lang="en">
  <head>
    <meta charset="UTF-8" />
    <meta name="viewport" content="width=device-width, initial-scale=1.0" />
    <title>Web-Bench Grid</title>
    <link href="./assets/tailwind.css" rel="stylesheet"/>
  </head>
  <body class="m-0">
    <div class="root grid grid-cols-[auto_1fr_auto] grid-rows-[auto_1fr_auto] h-screen">
      <div class="leftbar border-box bg-blue-200 p-4 col-start-1 row-start-1 row-span-3 w-64">Left Sidebar</div>
      <div class="header border-box bg-gray-200 p-4 border-b border-gray-400 col-start-2 row-start-1">Header Content</div>
      <div class="content border-box bg-gray-100 p-4 overflow-auto col-start-2 row-start-2">Main Content Area</div>
      <div class="footer border-box bg-gray-200 p-4 border-t border-gray-400 col-start-2 row-start-3">Footer Content</div>
      <div class="rightbar border-box bg-green-200 p-4 col-start-3 row-start-1 row-span-3 w-64">Right Sidebar</div>
    </div>
  </body>
</html>
```
\end{lstlisting}

\paragraph{Error} \ \\
\begin{center}
\includegraphics[width=0.9\linewidth]{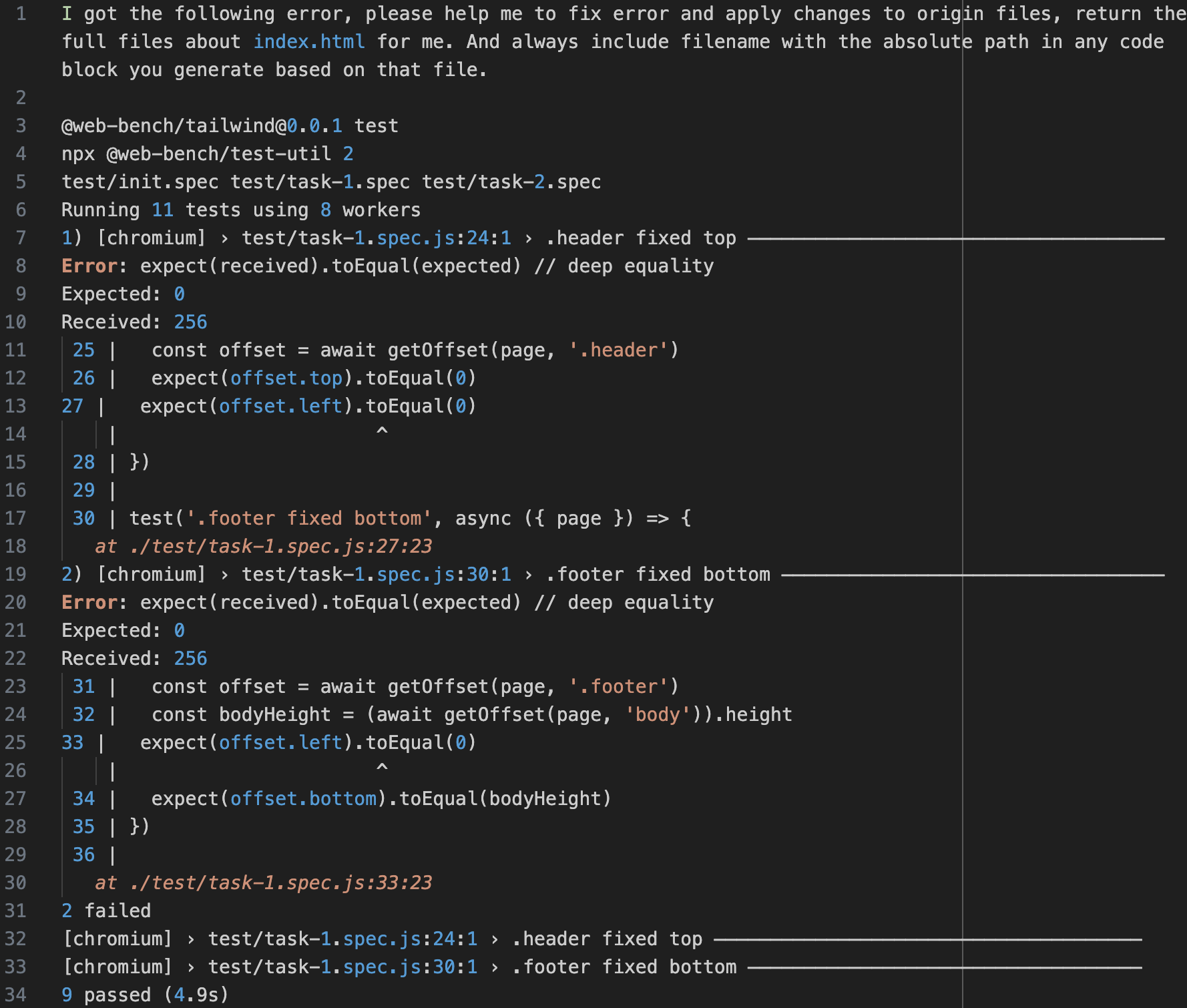}
\end{center}

\paragraph{Prompt} \ \\
Get the prompt that will be sent to LLM.
\begin{lstlisting}[style=json]
[
{
"role": "system",
"content": "\n# Rules\nWhen generating new code:\n1. Always produce a single code block.,2. Never separate the code into multiple code blocks.,3. Only include the code that is being added.,4. No explanation, no issue, only code.,5. Never omit any code.,6. If the user submits a code block that contains a filename in the language specifier, always include the filename in any code block you generate based on that file. The filename should be on the next line as the language specifier in your code block.,7. Don't repeat filename in code block\n\nAlways follow these guidelines when generating code responses.\n\n# Example\n\nHere is an example of response:\n\n<example>\n ```html\nfile_a.html \n<div>file_a</div>\n```\n```typescript \nsub_dir/file_b.ts\nconsole.log(\"file_b\") \n```\n</example>\n\nHere are some error examples of response:\n\n1. repeated filenames\n<example>\n```javascript\n index.js\nindex.js\n window.addEventListener('DOMContentLoaded', () => {\n    console.log('Dark mode page loaded');\n});\n```\n</example>\n\n2. without filename\n<example>\n```javascript\n window.addEventListener('DOMContentLoaded', () => {\n    console.log('Dark mode page loaded');\n});\n```\n</example>\n"
},
{
"role": "user",
"content": "```index.html\n<!DOCTYPE html>\n<html lang=\"en\">\n  <head>\n    <meta charset=\"UTF-8\" />\n    <meta name=\"viewport\" content=\"width=device-width, initial-scale=1.0\" />\n    <title>Web-Bench Grid</title>\n    <link href=\"./assets/tailwind.css\" rel=\"stylesheet\"/>\n  </head>\n  <body class=\"m-0\">\n    <div class=\"root grid grid-rows-[auto_1fr_auto] h-screen\">\n      <div class=\"header border-box bg-gray-200 p-4 border-b border-gray-400\">Header Content</div>\n      <div class=\"content border-box bg-gray-100 p-4 overflow-auto\">Main Content Area</div>\n      <div class=\"footer border-box bg-gray-200 p-4 border-t border-gray-400\">Footer Content</div>\n    </div>\n  </body>\n</html>\n```\nadd divs(class 'leftbar', 'rightbar') with arbitrary text in '.root' element. leftbar (border-box) is fixed at the left of '.root'; rightbar (border-box) is fixed at the right of '.root'; content occupies the remaining '.root' space. USE tailwind grid only. \n Do not compress the original code in file and return full file.\nI only want the returned results to contain code, without any explanations."
}
]
\end{lstlisting}
\paragraph{2. Request LLM}
As shown in Figure \ref{tab:model-parameters}. Requests the LLM's OpenAPI based on the selected model and returns the response. In this step, parameters such as temperature, maxToken, contextLength are obtained from the model provider. 
\paragraph{3. Extract Files}
Parses the LLM's response to extract the generated files. 

LLM response text is:

\begin{lstlisting}[style=html]
```html
index.html
<!DOCTYPE html>
<html lang="en">
  <head>
    <meta charset="UTF-8" />
    <meta name="viewport" content="width=device-width, initial-scale=1.0" />
    <title>Web-Bench Grid</title>
    <link href="./assets/tailwind.css" rel="stylesheet"/>
  </head>
  <body class="m-0">
    <div class="root grid grid-rows-[auto_1fr_auto] grid-cols-[auto_1fr_auto] h-screen">
      <div class="leftbar border-box bg-gray-300 p-4 col-span-1 row-span-2">This is the leftbar</div>
      <div class="header border-box bg-gray-200 p-4 col-span-2 row-span-1 sticky top-0 left-0">This is the header</div>
      <div class="rightbar border-box bg-gray-300 p-4 col-span-1 row-span-2">This is the rightbar</div>
      <div class="content border-box bg-gray-100 p-4 col-span-1 row-span-1">This is the content</div>
      <div class="footer border-box bg-gray-200 p-4 col-span-3 row-span-1 sticky bottom-0 left-0">This is the footer</div>
    </div>
  </body>
</html>
```
\end{lstlisting}

The generated result is:
\begin{lstlisting}[style=json]
{
"files": {
"index.html": "<!DOCTYPE html>\n<html lang=\"en\">\n  <head>\n    <meta charset=\"UTF-8\" />\n    <meta name=\"viewport\" content=\"width=device-width, initial-scale=1.0\" />\n    <title>Web-Bench Grid</title>\n    <link href=\"./assets/tailwind.css\" rel=\"stylesheet\"/>\n  </head>\n  <body class=\"m-0\">\n    <div class=\"root grid grid-rows-[auto_1fr_auto] grid-cols-[auto_1fr_auto] h-screen\">\n      <div class=\"leftbar border-box bg-gray-300 p-4 col-span-1 row-span-2\">This is the leftbar</div>\n      <div class=\"header border-box bg-gray-200 p-4 col-span-2 row-span-1 sticky top-0 left-0\">This is the header</div>\n      <div class=\"rightbar border-box bg-gray-300 p-4 col-span-1 row-span-2\">This is the rightbar</div>\n      <div class=\"content border-box bg-gray-100 p-4 col-span-1 row-span-1\">This is the content</div>\n      <div class=\"footer border-box bg-gray-200 p-4 col-span-3 row-span-1 sticky bottom-0 left-0\">This is the footer</div>\n    </div>\n  </body>\n</html>"
}
}
\end{lstlisting} 

\section{Appendix: Evaluator Tools}
\label{appendix:c}
\subsection{Evaluator}
\paragraph{Set Up} \href{https://github.com/bytedance/Web-Bench}{Web-Bench} uses Docker for reproducible evaluations. Refer to the Docker setup guide for instructions on installing Docker on your machine. To run Web-Bench from source, follow these steps:

\paragraph{1. Init} \ \\
\begin{lstlisting}[style=bash]
git clone https://github.com/bytedance/Web-Bench.git
cd Web-Bench
\end{lstlisting}

\paragraph{2. Config} config files 'apps/eval/src/config.json5' must be created locally. Configuration details can be found in Table \ref{tab:eval-config}.

\begin{table*}[ht]
  \centering
  \caption{Evaluator Configuration}
  \begin{tabular}{lp{3cm}cp{8cm}}
  \toprule
  Parameter & Type & Default & Description \\
  \midrule
  projects & string[] & [] & If empty, all projects will pass.  Value is like "\@web-bench/calculator". \\
  agentMode & "local" $|$ "http" & "local" &  \\
  agentEndPoint & string & "" & When agentMode is set to "http", set http API for network requests. \\
  models & string[] & [] & 'models' field  in apps/eval/src/model.json \\
  maxdop & number & 30 & max degree of parallelism \\
  logLevel & "info" $|$ "warn" $|$ "debug" $|$ "error" & "info" &  \\
  httpLimit & number & 10 & When agentMode is set to "http", maximum concurrent requests \\
  projects & string[] & [] & If empty, all projects will pass.  Value is like "\@web-bench/calculator". \\
  agentMode & "local" $|$ "http" & "local" &  \\
  agentEndPoint & string & "" & When agentMode is set to "http", set http API for network requests. \\
  \bottomrule 
  \end{tabular}
  \label{tab:eval-config}
\end{table*}

\paragraph{Use HTTP-Agent} \ \\
\begin{lstlisting}[style=json]
  {
    "agentMode": "http",
    "agentEndPoint": "http://example.com/path/to/agent"
  }
\end{lstlisting}

\paragraph{Use Local-Agent}
It is actually the Web-Agent introduced by this paper \ref{appendix:b}.
\begin{lstlisting}[style=json]
  {
    "agentMode": "local",
    "models": ["claude-3-5-sonnet-20241022", "openai/gpt-4o"]
  }
\end{lstlisting}
and the file 'apps/eval/.env' must be created locally
\begin{lstlisting}[style=bash]
  ANTHROPIC_API_KEY=xxx
  OPENROUTER_API_KEY=xxx
  # Comment
  # OPENAI_API_KEY=xxx
\end{lstlisting}

'XXX\_API\_KEY' is used by 'apps/eval/src/model.json'\\
\begin{lstlisting}[style=json]
  {
    "models": [
      {
        "model": "claude-3-5-sonnet-20241022",
        "apiKey": "{{ANTHROPIC_API_KEY}}"
      },
      {
        "model": "openai/gpt-4o",
        "apiKey": "{{OPENROUTER_API_KEY}}"
      }
    ]
  }
\end{lstlisting}

\paragraph{3. Build} \ \\
\begin{lstlisting}[style=bash]
docker build -f ./start.dockerfile -t web-bench .
\end{lstlisting}

\paragraph{Usage}  \ \\
\begin{lstlisting}[style=bash]
docker run web-bench
\end{lstlisting}
The results will be saved in 'apps/eval/report'.

\paragraph{Q \& A} 
\paragraph{Difference between Local-Agent and HTTP-Agent} 
\ \\
- Local-Agent: Local-Agent has the basic capability to interact with LLM. It can specify the corresponding model in the 'apps/eval/src/model.json'. Local-Agent is actually the Web-Agent introduced by this paper \ref{appendix:b}.\\
- HTTP-Agent:  Through HTTP-Agent, it calls the configured agentEndPoint to send a request to the custom Agent.

\paragraph{Add new model for evaluation} 
\ \\
1. For models deployed on OpenRouter, use the native OpenRouter provider with the following configuration:
\begin{lstlisting}[style=json]
{
  "title": "anthropic/claude-3-opus",
  "provider": "openrouter",
  "model": "anthropic/claude-3-opus",
  "apiBase": "https://openrouter.ai/api/v1",
  "apiKey": "{{OPENROUTER_API_KEY}}"
}
\end{lstlisting}

\noindent
2. If existing providers do not meet your requirements, you can evaluate specific models by creating a new Provider. This is achieved by extending the BaseLLM:
\begin{lstlisting}[style=js]
export abstract class BaseLLM {
  abstract provider: string
  abstract option: LLMOption
  info: Model
  abstract chat(
    compiledMessages: ChatMessage[],
    originOptions: CompletionOptions
  ): Promise<{
    request: string
    error?: string
    response: string
  }>
}
\end{lstlisting}

1. option – define parameters for LLM requests:
\begin{lstlisting}[style=js]
    export interface LLMOption {
      contextLength: number
      maxTokens: number
      temperature?: number
      apiBase: string
    }
\end{lstlisting}

2. info – model metadata in apps/eval/src/model.json.

3. chat – custom request method that returns the generated text from the LLM.

\paragraph{Use custom agent for evaluation} Web-Bench supports custom agent by built-in HTTP-Agent. During the "Call Agent" step \ref{sec:evaluator-workflow}, the HTTP-Agent will:

1. Pass the Evaluator's context to your CustomAgent.

2. Return the CustomAgent's response to the Evaluator without modifications.

The request and response formats can be found in the next section. \ref{appendix:c-agent}

\subsection{Agent Data Format}
\label{appendix:c-agent}
\paragraph{HTTP Request}
\paragraph{HTTP Request Method}  \ \\
POST
\paragraph{HTTP Request Body} \ \\
\begin{lstlisting}[style=js]
export interface AgentRequest {
  type: "normal" | "init"
  task: string
  // Code files, key is filePath, value is fileContent
  files?: Record<string, string>
  // Error context
  error?: string
}
\end{lstlisting}

\paragraph{HTTP Response}
\paragraph{HTTP Response Status} \ \\
- 200, Successful\\
- 500, Failed
\paragraph{HTTP Response Body} \ \\
\begin{lstlisting}[style=js]
export interface AgentResponse {
  // Code files, key is filePath, value is fileContent
  files: Record<string, string>
}
\end{lstlisting}

\paragraph{Example - Init Task}
\paragraph{Request} \ \\
- Method: POST\\
- Body:
\begin{lstlisting}[style=json]
{
"type": "init",
"task": " generate a calculator in a single HTML file. the first row should be an input element with id 'display'; the next 4 rows should contain buttons with digits from '0' to '9' and operators including '+-*/=.'; the last row should have a 'Clear' button. display 'Error' when catching exception or getting undefined value during calculating. And add the html file filename after code block. The filename should be on the next line as the language specifier in your code block. the filename is \"index.html\""
}
\end{lstlisting}

\paragraph{Response} \ \\
- Status: 200\\
- Body:
\begin{lstlisting}[style=json]
{
"files":{
  "index.html": "<!DOCTYPE html>\n<html lang=\"en\">\n<head>\n    <meta charset=\"UTF-8\">\n    <meta name=\"viewport\" content=\"width=device-width, initial-scale=1.0\">\n    <title>Calculator</title>\n    <style> body {}\n  .calculator {} </style>\n</head>\n <body></body>\n</html>"
}
}
\end{lstlisting}

\paragraph{Example - Normal Task with Error}
\paragraph{Request} \ \\
- Method: POST\\
- Body:\\
\indent - Note: All file contents of the project will be transferred. Benchmark projects are relatively simple at present (mostly single files, a few single-digit files), and in the future, when the project structure becomes more complex, it will support shared folders to reduce the amount of data transferred.
\begin{lstlisting}[style=json]
{
"type": "normal",
"task": "add button sqrt with text 'sqrt' at the right of button clear; click it to calculate result using display content directly",
"files": {
  "index.html": "<!DOCTYPE html>\n<html lang=\"en\">\n<head>\n    <meta charset=\"UTF-8\">\n    <meta name=\"viewport\" content=\"width=device-width, initial-scale=1.0\">\n    <title>Calculator</title>\n    <style></style>\n</head>\n <body></body>\n</html>\n\n",
  "index.css": "body { }\n\n.calculator { }",
  "index.js": "let displayValue = '' // more"
},
"error": "\t\t\t\t\tAn error occurred during the execution of the unit test. "
}
\end{lstlisting}

\paragraph{Response} \ \\
- Status: 200\\
- Body:
\begin{lstlisting}[style=json]
{
"files": {
  "index.html": "<!DOCTYPE html>\n<html lang=\"en\">\n<head>\n    <meta charset=\"UTF-8\">\n    <meta name=\"viewport\" content=\"width=device-width, initial-scale=1.0\">\n    <title>Calculator</title>\n    <style></style>\n</head>\n <body></body>\n</html>\n\n",
  "index.css": "body {}",
  "index.js": "let displayValue = '' // more",
}
}
\end{lstlisting}

\section{Appendix: Calibrate Project}
\label{appendix:d}
\subsection{Workflow}
Use rush eval to Calibrate a project. That is, check for design issues in the project: 

1. tasks.jsonl issues, such as ambiguous descriptions, vulnerabilities, or obvious conflicts with test cases. 

2. test/ cases issues, such as not rigorous enough or too strict.

The current path is '/path/to/projects/dom'.

\noindent
1. rush eval with 'config.json5'. \\
\begin{lstlisting}[style=js]
{  
  projects: ['@web-bench/selector'],
  models: ['claude-3-5-sonnet-20241022'],
}
\end{lstlisting}
2. IF: task-20 passed, end\\
3. ELSE: use eval project files to replace 'src/':\\
\begin{lstlisting}[style=bash]
mv src src-20 # your manual codes
cp -r eval/eval-2024xx/claude-3-5-xx/task-5-2 src
\end{lstlisting}
4. IF: Modify 'src/' or test/ codes\\
\indent 1. IF: All tests passed in 'src/' and src-20/\\
\indent 2. THEN: rush eval with new config.json5\\
\begin{lstlisting}[style=js]
  {
    projects: ['@web-bench/selector'],
    models: ['claude-3-5-sonnet-20241022'],
    startTask: 'task-6',
  }
\end{lstlisting}
  3. goto Step-2
5. IF: Modify tasks.jsonl, task-n is the smallest index of modified tasks\\
\indent 1. THEN:  use successful task-{n-1} eval project files to replace 'src/':\\
\begin{lstlisting}[style=bash]
  rm -rf src
  cp -r eval/eval-2024xx/claude-3-5-xx/task-{n-1} src
\end{lstlisting}
\ \  2. THEN:rush eval with new config.json5\\
\begin{lstlisting}[style=js]
  {
    packageNames: ['@web-bench/selector'],
    models: ['claude-3-5-sonnet-20241022'],
    startTask: 'task-n',
  }
\end{lstlisting}
\ \  3. goto Step 2\\
6. Create a doc from template \ref{appendix:d-template}.

\subsection{Template}
\label{appendix:d-template}

Calibrate with 'claude-3-5-sonnet-20241022':\\
\includegraphics[width=1\linewidth]{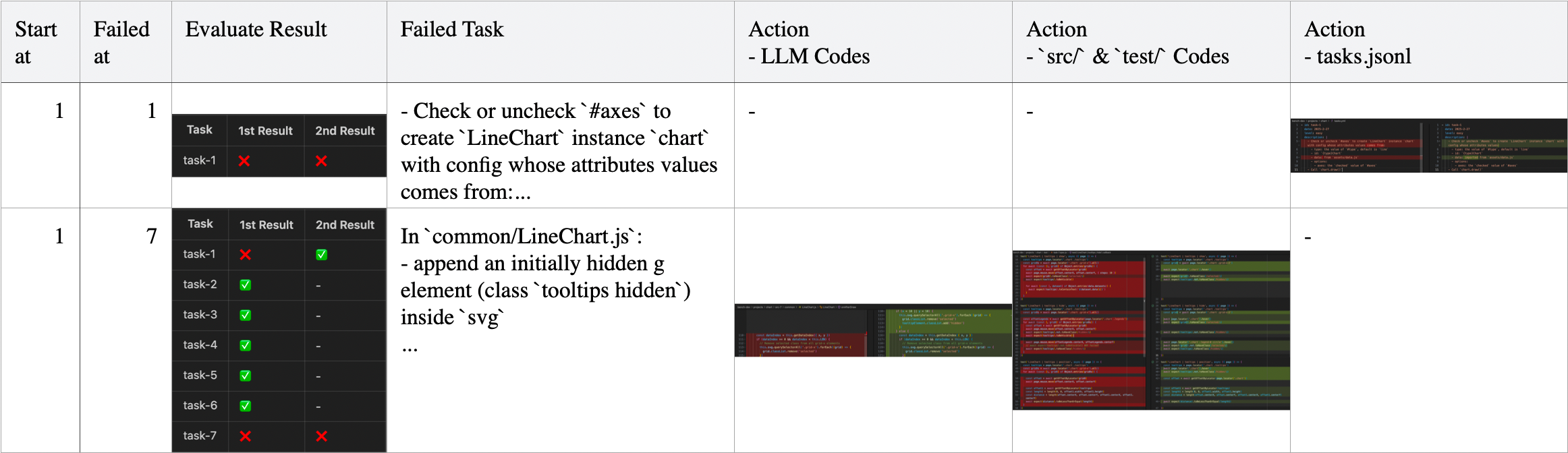}

\section{Appendix: Evaluation Models}
\label{appendix:e}
See Table \ref{tab:models}.

\begin{table*}[htbp!]
  \centering
  \caption{Evaluation Models}
  \begin{tabular}{llcccl}
  \toprule
  Model & Provider & Pass@1 & Pass@2 & Error@1 & Type \\
  \midrule
  gpt-4.1 & OpenAI & 21.09\% & 25.11\% & 6.52\% & Closed \\
  gpt-4.1-mini & OpenAI & 20.76\% & 23.70\% & 6.85\% & Closed \\
  gpt-4o & OpenAI & 17.17\% & 23.80\% & 8.15\% & Closed \\
  o4-mini & OpenAI & 13.26\% & 22.93\% & 9.89\% & Closed \\
  o1 & OpenAI & 10.43\% & 12.39\% & 9.89\% & Closed \\
  o3-mini & OpenAI & 9.13\% & 14.24\% & 7.93\% & Closed \\
  gpt-4o-mini & OpenAI & 8.48\% & 13.04\% & 7.28\% & Closed \\
  gpt-4.1-nano & OpenAI & 7.07\% & 12.28\% & 6.96\% & Closed \\
  claude-3-7-sonnet-20250219-thinking & Anthropic & 25.11\% & 35.33\% & 9.02\% & Closed \\
  claude-3-5-sonnet-20241022 & Anthropic & 23.04\% & 32.39\% & 9.46\% & Closed \\
  claude-3-7-sonnet-20250219 & Anthropic & 22.50\% & 30.98\% & 8.26\% & Closed \\
  claude-3-5-sonnet-20240620 & Anthropic & 21.96\% & 30.33\% & 8.80\% & Closed \\
  claude-3-5-haiku-20241022 & Anthropic & 15.43\% & 21.74\% & 7.61\% & Closed \\
  deepseek-v3-0324 & DeepSeek & 17.07\% & 23.59\% & 7.93\% & Open \\
  deepseek-coder-v2 & DeepSeek & 16.74\% & 23.15\% & 7.83\% & Open \\
  deepseek-r1 & DeepSeek & 14.46\% & 26.20\% & 9.89\% & Open \\
  gemini-2.5-pro & Google & 15.67\% & 24.02\% & 8.04\% & Closed \\
  gemini-2.0-flash & Google & 15.33\% & 20.87\% & 7.07\% & Closed \\
  gemini-2.0-flash-thinking & Google & 14.89\% & 19.24\% & 6.96\% & Closed \\
  gemini-pro-1.5 & Google & 14.78\% & 20.87\% & 7.28\% & Closed \\
  gemini-flash-1.5 & Google & 12.83\% & 17.07\% & 7.07\% & Closed \\
  gemma-3-27b & Google & 9.89\% & 11.85\% & 6.20\% & Open \\
  llama-4 Maverick & Meta & 15.98\% & 20.87\% & 7.07\% & Open \\
  llama-3.3 & Meta & 6.63\% & 9.57\% & 6.20\% & Open \\
  llama-4 Scout & Meta & 5.00\% & 7.72\% & 6.63\% & Open \\
  qwen-max-2025-01-25 & QWen & 15.87\% & 19.02\% & 6.30\% & Open \\
  qwen-plus-2025-01-25 & QWen & 11.85\% & 15.11\% & 6.20\% & Open \\
  qwen-2.5-72b-instruct & QWen & 10.54\% & 13.70\% & 6.63\% & Open \\
  qwen-turbo-2024-11-01 & QWen & 2.61\% & 5.11\% & 6.85\% & Open \\
  mistral-large-2411 & MistralAI & 13.04\% & 18.70\% & 7.28\% & Open \\
  grok-2-1212 & X-AI & 11.30\% & 17.17\% & 7.93\% & Closed \\
  doubao-pro-1.5-thinking & Doubao & 20.11\% & 30.22\% & 10.56\% & Closed \\
  doubao-pro-1.5-32k & Doubao & 16.63\% & 22.93\% & 7.28\% & Closed \\
  doubao-pro-1.5-32k-lite & Doubao & 3.48\% & 5.98\% & 6.41\% & Closed \\
  GLM-4-0414 & Thudm & 7.50\% & 9.02\% & 6.30\% & Open \\
  step-fun-2-16k & StepFun & 13.70\% & 15.87\% & 6.09\% & Closed \\
  sense-chat-5 & SenseTime & 8.48\% & 12.72\% & 6.85\% & Closed \\
  360-gpt2-o1 & 360 & 8.26\% & 14.46\% & 7.39\% & Closed \\
  minimax-text & MiniMax & 8.48\% & 10.76\% & 6.20\% & Open \\
  moonshot-kimi-latest & MoonshotAI & 5.22\% & 11.85\% & 10.22\% & Open \\
  \bottomrule 
  \end{tabular}
  \label{tab:models}
\end{table*}

\end{document}